\documentclass[11pt,a4paper]{article}
\usepackage[hyperref]{acl2020}
\usepackage{times}  %Required
\usepackage{url}  %Required
\usepackage{subcaption}
\usepackage{graphicx}  %Required
\usepackage{multirow}
\usepackage{natbib}
\usepackage{boxedminipage}

\newcommand{\Eq}[1]{\mbox{Eq.~(\ref{eq:#1})}}
\newcommand{\Sec}[1]{{Section~\ref{section:#1}}}
\newcommand{\Tab}[1]{{Table~\ref{table:#1}}}
\newcommand{\Fig}[1]{{Figure~\ref{figure:#1}}}
\newcommand{\sig}{{$^{\dagger}$}}
\newcommand{\unidir}[2]{{#1$\rightarrow$#2}}
\newcommand{\App}[1]{{Appendix~\ref{appendix:#1}}}

\aclfinalcopy

\begin{document}

\title{Recurrent Stacking of Layers in Neural Networks:\\ An Application to Neural Machine Translation}

\author{Raj Dabre \qquad Atsushi Fujita\\
National Institute of Information and Communications Technology \\
3-5 Hikaridai, Seika-cho, Soraku-gun, Kyoto, 619-0289, Japan\\
\tt{firstname.lastname@nict.go.jp}}

\maketitle

\begin{abstract}
In deep neural network modeling, the most common practice is to stack a number of recurrent, convolutional, or feed-forward layers in order to obtain high-quality continuous space representations which in turn improves the quality of the network's prediction. Conventionally, each layer in the stack has its own parameters which leads to a significant increase in the number of model parameters. In this paper, we propose to share parameters across all layers thereby leading to a recurrently stacked neural network model. We report on an extensive case study on neural machine translation (NMT), where we apply our proposed method to an encoder-decoder based neural network model, i.e., the Transformer model, and experiment with three Japanese--English translation datasets. We empirically demonstrate that the translation quality of a model that recurrently stacks a single layer 6 times, despite having significantly fewer parameters, approaches that of a model that stacks 6 layers where each layer has different parameters. We also explore the limits of recurrent stacking where we train extremely deep NMT models. This paper also examines the utility of our recurrently stacked model as a student model through transfer learning via leveraging pre-trained parameters and knowledge distillation, and shows that it compensates for the performance drops in translation quality that the direct training of recurrently stacked model brings. We also show how transfer learning helps in faster decoding on top of the already reduced number of parameters due to recurrent stacking.  Finally, we analyze the effects of recurrently stacked layers by visualizing the attentions of models that use recurrently stacked layers and models that do not.

\end{abstract}

\section{Introduction}
\label{section:introduction}

Deep learning approaches allow for end-to-end training of neural network models that transform inputs into desired outputs. These inputs and outputs are typically image, text, and/or audio. Most successful deep learning models involve stacking a number of layers that convert inputs into continuous space vectors and then into outputs. Typically, such layers consist of either recurrent (mostly for text and audio), convolutional (mostly for image), or self-attentional feed-forward (image or text) components.
It has been empirically demonstrated that stacking of layers leads to an improvement in performance, especially in high-resource scenarios, since it allows for better representation learning owing to additional parameters
and/or more multiplications of non-linear transformations. However, it also increases both space complexity, e.g., the number of parameters (or simply ``size'') of the model, and time complexity, e.g., the time consumed during training and inference, by a significant amount.

\begin{figure*}[t]
    \centering
    \begin{subfigure}[b]{0.3\textwidth}\centering
    \includegraphics*[scale=.45]{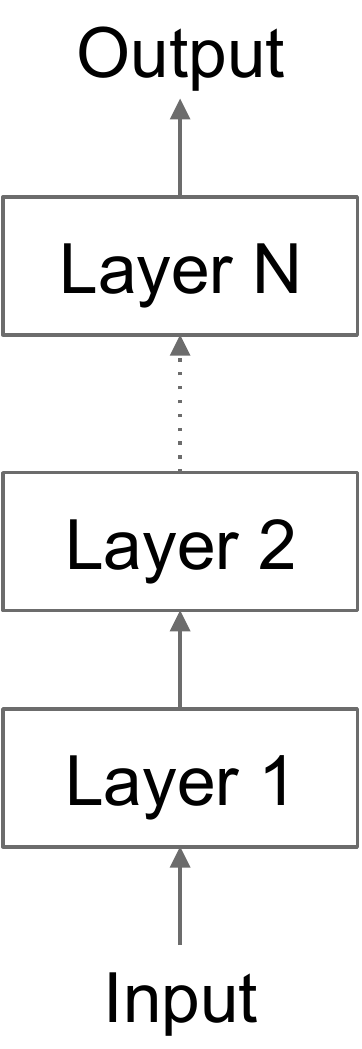}
    \caption{Vanilla model}
    \end{subfigure}
    \begin{subfigure}[b]{0.3\textwidth}\centering
    \includegraphics*[scale=.45]{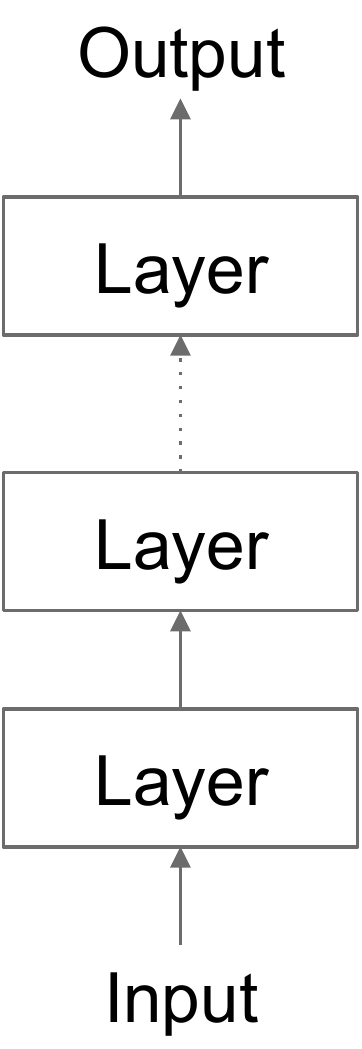}
    \caption{RS model (unfold)}
    \end{subfigure}
    \begin{subfigure}[b]{0.3\textwidth}\centering
    \includegraphics*[scale=.45]{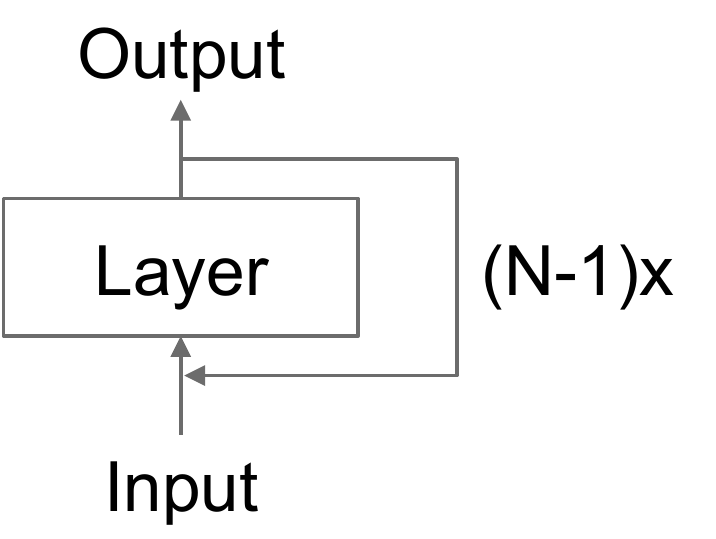}
    \caption{RS model (fold)}
    \end{subfigure}
    \caption{Comparison of vanilla and recurrently stacked (RS) models: (a) vanilla model has different sets of parameters in different layers, whereas (b) RS model uses the same parameters multiple times. (c) is a simplified version of (b).}
    \label{figure:vanilla-vs-RS}
\end{figure*}

This paper focuses on compact model architectures through simplification of existing architectures. Compact models are desirable in practical scenarios, especially when only machines with low memory are available for deployment.
The search for compact neural network models with performance comparable to their bulkier counterparts has led to the development of \emph{knowledge distillation} approaches \citep{DBLP:journals/corr/HintonVD15,kim-rush-2016-sequence,DBLP:journals/corr/FreitagAS17}. In this approach, teacher\footnote{In literature related to transfer learning by fine-tuning, teacher and student models are commonly known as parent and child models, respectively. Even though we consider that these terms are interchangeable, in this paper, we use the former terms in order to be consistent with teacher-student terminology used in knowledge distillation.} models with a large number of parameters are first trained and then student models with significantly fewer parameters are trained to mimic the teacher models. This mimicking takes place in the form of matching the probability distribution of teacher models. As a result, a single student model with a small number of parameters can imbibe the performance of a teacher model with a significantly larger number of parameters. Additionally, such a student model can also be taught to perform as well as the ensemble of larger teacher models which outperforms the individual teacher models.

As an alternative approach for model compression, in this paper, we propose \emph{recurrent stacking} (RS)\footnote{RS can mean ``recurrently stacked'' or ``recurrent stacking'' both of which have the same implication.} of layers. \Fig{vanilla-vs-RS} compares vanilla and RS models with $N$ layers.  Unlike vanilla models, where each of the $N$ stacked layers has its own parameters, RS models use the same parameters across all the $N$ stacked layers, significantly reducing the size of the model compared to the vanilla model with $N$ layers. Our approach was motivated by works on parameter sharing architectures \cite{TACL1081,DBLP:conf/acl/DongWHYW15,kaiser:17} where the same model parameters are reused for several diverse tasks without strong negative effects to the performance of those tasks. We rationalized that if the sharing parameters between tasks is not detrimental to those tasks' performance then sharing parameters between components of a model for a single task should not be detrimental to that task's performance either.
The concept of RS does not assume characteristics of specific neural network models and tasks, and thus it should be applicable to any neural network architectures in general.  In this paper, we empirically evaluate its utility, taking a case study of neural machine translation (NMT) \citep{DBLP:journals/corr/ChoMGBSB14,DBLP:journals/corr/BahdanauCB14:original}.  Following the recent advances in NMT, we work with the \textit{Transformer} architecture \citep{NIPS2017_7181}, an instance of encoder-decoder neural networks, which are prevalently used for sequence-to-sequence (seq2seq) modeling \citep{DBLP:journals/corr/SutskeverVL14}.
Through experiments on three different datasets, we clarify whether it is the additional parameters or the additional non-linear transformations that are responsible for improvement in performance as a result of stacking layers.

These experiments reveal that RS models always have better performance than the 1-layer vanilla models, i.e., the models with exactly the same number of parameters.  However, when compared to the 6-layer vanilla models, RS models tend to suffer from reduced performance.
To obtain compact models with a simple architecture but with comparable performance to vanilla models, we also investigate the utility of RS models as student models in a transfer learning setting.  More specifically, we train RS models through two types of transfer learning techniques: a simple layer transfer followed by fine-tuning \citep{DBLP:conf/emnlp/ZophYMK16:original} and sequence-level knowledge distillation \citep{kim-rush-2016-sequence}.
For the first approach, we initialize the layer of an RS model with a particular layer from a pre-trained vanilla model and then perform fine-tuning of the parameters. For the second, we generate pseudo-training data with the pre-trained vanilla model and then train the RS model on the resultant data. The former is an explicit transfer learning method whereas the latter is an implicit transfer learning method.
Through additional experiments, we confirm that transfer learning methods can compensate for the performance drops that the direct training of the RS models suffer from.

This paper is an extended version of our previous work on RS layer models \cite{DBLP:conf/aaai/DabreF19} with additional novel contents. The rest of this paper is organized as follows. In \Sec{relwork}, we review existing methods that focus on the sizes of deep neural network models in general.
Then, the following three contributions of this paper are presented.
\begin{description}
    \item[Sections~\ref{section:proposed} and \ref{section:experiments}:] We propose RS of layers and empirically evaluate it using NMT as an application. (our previous work + minor new work + refinements)
    \item[\Sec{transfer_learning}:] We also propose to train RS models through transfer learning to bridge the performance gaps between RS and vanilla models. (completely new work)
    \item[\Sec{analysis}:] By visualizing the attentions across different layers of various RS models, we unveil the adaptive power of their attention mechanisms. (our previous work + refinements)
\end{description}
Finally, \Sec{conclusion} concludes this paper, mentioning future research directions. 

\section{Related Work}
\label{section:relwork}

There are three major bodies of work that deal with reducing the sizes of deep neural network models: (a)~parameter sharing, (b)~transfer learning, and (c)~implementation tricks.

The simplest way to reduce the number of parameters in a neural network model is to find parts that can be shared.  This \textbf{parameter sharing} has been examined in a wide variety of scenarios.
The work on zero-shot NMT \citep{TACL1081} revealed that a single encoder-decoder NMT model can host multiple language pairs without an appreciable loss in translation quality,
thereby avoiding the need to train separate models per translation direction. For languages that share orthographies \citep{sennrich-haddow-birch:2016:P16-12} or have orthographies that can be mapped to a common script \cite{kunchukuttan-etal-2018-leveraging}, using a shared embedding layer can help reduce the model size significantly, also enabling cognate sharing.
Universal Transformer \citep{univtrans} shows that feeding the output of the multi-layer encoder (and decoder) to itself repeatedly leads to an improvement in quality for English-to-German translation. The idea of RS is similar to this in the sense that both use the same parameters recurrently. However, whereas \citet{univtrans} have focused on improving the state-of-the-art, the aim of RS is to obtain neural network models with significantly fewer parameters.  Eventually, the RS models have the same size as that of a 1-layer model.
Another approach is to share the parameters between the encoder and the decoder \cite{DBLP:conf/aaai/XiaHTTHQ19,DBLP:conf/aaai/DabreF19}.
We consider that this approach is orthogonal to our RS and will examine their combination in our future work.

Another common way to reduce the size of a neural network model is \textbf{transfer learning} through knowledge distillation \citep{DBLP:journals/corr/HintonVD15,kim-rush-2016-sequence,DBLP:journals/corr/FreitagAS17}, which transfers the knowledge learned by the larger teacher model(s) into a smaller student model.
It requires training one or more teacher models and thus is time-consuming.
In this paper, we examine the combination of RS and knowledge distillation to yield compact neural network models without performance loss. Additionally, we examine transfer learning by fine-tuning pre-trained models \cite{DBLP:conf/emnlp/ZophYMK16:original} unlike most work.

\begin{table*}[t]
\centering
\small
\begin{tabular}{c|c|c|c|c|c}
\hline
Dataset & Train & Dev & Test & Vocab & Training iteration \\\hline
  GCP & 400k & 2,000 & 2,000 & 16k & 60k/120k\\
  KFTT & 440k & 1,166 & 1,160 & 8k & 160k\\
  ASPEC & 1.50M & 1,790 & 1,812 & 32k & 400k\\\hline
\end{tabular}
\caption{Datasets (number of sentence pairs) and model settings.}
\label{table:datasets}
\end{table*}

Finally, the third orthogonal body of work involves \textbf{implementation tricks}. Pruning of pre-trained models \cite{see-etal-2016-compression} makes it possible to discard around 80\% of the smallest weights of a model without deterioration in performance. Such models do require re-training and hyper-parameter tuning after pruning to avoid loss in performance. Currently, most deep learning implementations use 32-bit floating point representations, but 16-bit floating point representations can be an alternative without any loss in performance \cite{pmlr-v37-gupta15,ott-etal-2018-scaling}.
There is an even more aggressive work that uses binarization \cite{Courbariaux:16} as a way to compress models. These methods can naturally be used with the RS approach and lead to a speed-up in decoding due to faster computations.  The RS approach should be combined with binary code prediction method \cite{oda-etal-2017-neural}, which specifically intends to speed up the softmax layer of neural networks. On a related topic, compact models are usually faster to decode and work on quantization \cite{Lin:2016:FPQ:3045390.3045690} and average-attention networks \citep{DBLP:conf/acl/XiongZS18} address the topic of fast decoding in detail.

\section{Recurrent Stacking of Layers}
\label{section:proposed}

In this paper, we propose to share the parameters across layers by \emph{recurrent stacking} (RS), as illustrated in \Fig{vanilla-vs-RS}.

Assume that $X$ is the input to the neural network model consisting of $N$ layers. Let $Y$ be the output of the topmost layer, $L_{i}$ be the $i$-th layer, and $L_{i}(\cdot)$ indicates that $L_{i}$ transforms the argument into a hidden representation. \Eq{vanilla-eq} shows the hidden layer computation of a 6-layer vanilla model. In contrast, as shown in \Eq{rs-eq}, a 6-layer RS model uses the same layer 6 times.
\begin{eqnarray}
    Y &= &L_{6}(L_{5}(L_{4}(L_{3}(L_{2}(L_{1}(X))))))
    \label{eq:vanilla-eq}\\
    Y &= &L_{1}(L_{1}(L_{1}(L_{1}(L_{1}(L_{1}(X))))))
    \label{eq:rs-eq}
\end{eqnarray}

This parameter sharing results in a single-layer deep neural network model, massively reducing the model size. During training, RS of layers enforces the model to revise the hidden states and leads to more complex representations that should help in improving translation quality.

Let us consider an application of RS to encoder-decoder models consisting of (a)~an encoder comprising of an embedding layer for the input and $N$ stacked transformation layers, and (b)~a decoder comprising of an embedding layer, $M$ stacked transformation layers, and a softmax layer for the output, where $M$ is often set to $N$.
RS is applied to each of the encoder and the decoder. Beside this approach, there are two well-known orthogonal approaches for sharing parameters of encoder-decoder models. One involves using shared input-output vocabulary; the same parameters are used for embeddings of both the encoder and the decoder and the softmax layer of the decoder. The other is to share the parameters between the encoder and the decoder \citep{DBLP:conf/aaai/XiaHTTHQ19}.

\section{Machine Translation Experiments}
\label{section:experiments}

We empirically evaluated the utility of our proposed method through its application to NMT, using the Transformer model \citep{NIPS2017_7181}.
Even though our method is independent of implementation and model as formulated in \Sec{proposed}, we chose the Transformer model, since it is the current state-of-the-art NMT model.
In the case of the Transformer models, each layer consists of self-attention, cross-attention (decoder only), and feed-forward sub-layers with layer normalization and residual connections for each sub-layer.
To clarify the contribution of our RS method, we trained and evaluated the following two types of NMT models, where the same numbers of encoder and decoder layers were used.
\begin{description}
    \item[(a) Vanilla NMT:] $k$-layer ($k\in\{1,2,6\}$) models without any shared parameters across layers.
    \item[(b) RS-NMT:] $k$-layer ($k\in\{1,2,3,4,5,6\}$) models with parameters shared across all layers. Note that the 1-layer model is identical to the 1-layer vanilla NMT model.
\end{description}

\begin{table*}[t]
\centering
\small
\begin{tabular}{c|c|c|c|c|c|c}
\hline
\multirow{2}{*}{Model Type}  & \multicolumn{2}{c|}{GCP} & \multicolumn{2}{c|}{KFTT} & \multicolumn{2}{c}{ASPEC} \\ \cline{2-7} 
  & \unidir{Ja}{En} & \unidir{En}{Ja} & \unidir{Ja}{En} & \unidir{En}{Ja} & \unidir{Ja}{En} & \unidir{En}{Ja} \\ \hline
\multirow{2}{*}{\begin{tabular}[c]{@{}c@{}}RS-NMT and\\ 1-layer vanilla NMT\end{tabular}}& \multirow{2}{*}{30.4M} & \multirow{2}{*}{33.2M} & \multirow{2}{*}{20.2M} & \multirow{2}{*}{20.3M} & \multirow{2}{*}{57.7M} & \multirow{2}{*}{57.7M} \\
  &  &  &  &  &  &  \\
\hline
2-layer vanilla NMT  & 39.5M & 40.4M & 28.4M & 28.6M & 65.0M & 65.2M \\ \hline
6-layer vanilla NMT  & 69.8M & 70.2M & 57.7M & 57.9M & 94.4M & 96.2M \\ \hline

\end{tabular}
\caption{The numbers of parameters for different NMT models that we trained.}
\label{table:model-parameters}
\end{table*}

\begin{table*}[t]
\centering
\small
\begin{tabular}{c|c|c|c|c|c|c}
\hline
\multirow{2}{*}{Model} & \multicolumn{2}{c|}{GCP} & \multicolumn{2}{c|}{KFTT} & \multicolumn{2}{c}{ASPEC} \\ \cline{2-7} 
 & \unidir{Ja}{En} & \unidir{En}{Ja} & \unidir{Ja}{En} & \unidir{En}{Ja} & \unidir{Ja}{En} & \unidir{En}{Ja} \\ \hline

\multicolumn{7}{l}{\textbf{(a) Vanilla NMT}}\\\hline
1 & 21.95 & 23.89 & 21.64 & 25.00 & 23.28 & 32.19 \\
2 & 24.23\sig & 25.62 & 24.14 & 30.05 & 28.06 & 38.91 \\
6 & \textbf{24.67} & \textbf{26.22} & \textbf{27.19} & \textbf{32.72} & \textbf{28.77} & \textbf{41.32} \\ \hline

\multicolumn{7}{l}{\textbf{(b) RS-NMT}}\\\hline
1 & 21.95 & 23.89 & 21.64 & 25.00 & 23.28 & 32.19 \\
2 & 23.24 & 24.47 & 24.50 & 28.53 & 27.84 & 38.54 \\
3 & 23.42 & 25.02 & 25.84 & 29.90 & 28.05 & 39.26 \\
4 & 24.33 & 25.28 & 26.23 & 30.36 & 28.08 & 39.31 \\
5 & 23.95 & 25.38 & 26.42 & 30.78 & 28.02 & 38.86 \\
6 & 24.36\sig & 25.84\sig & 26.51 & 30.83 & 27.20 & 40.04 \\ \hline

\end{tabular}
\caption{BLEU scores obtained in our experiments. The definition of numbers in the leftmost column are as follows: (a)~the number of different encoder and decoder layers and (b)~the number of recurrence. Scores in bold are the ones with highest BLEU score in each translation task, whereas scores marked with ``\sig'' are the ones that are not statistically significantly (bootstrap re-sampling with $p<0.05$) worse compared to the corresponding 6-layer vanilla model.}
\label{table:direct-results}
\end{table*}

\subsection{Settings}

We used three different datasets for Japanese--English translation for both directions (\unidir{Ja}{En} and \unidir{En}{Ja}):
the Global Communication Plan (GCP) corpus\footnote{The splits provided by \citet{W18-2707}.} \cite{IMAMURA18.104}, the Kyoto free translation task (KFTT) corpus,\footnote{\url{http://www.phontron.com/kftt}} 
and the Asian Scientific Paper Excerpt Corpus (ASPEC)\footnote{\url{http://orchid.kuee.kyoto-u.ac.jp/ASPEC/} We used the cleaner half of this corpus.} \citep{NAKAZAWA16.621}.
\Tab{datasets} gives the number of parallel sentences for all the datasets.
We tokenized the Japanese sentences in the KFTT and ASPEC corpora using the JUMAN morphological analyzer.\footnote{\url{http://nlp.ist.i.kyoto-u.ac.jp/EN/index.php?JUMAN}} We tokenized and lowercased the English sentences of these corpora using the \textit{tokenizer.perl} and \textit{lowercase.perl} scripts in Moses.\footnote{\url{https://github.com/moses-smt/mosesdecoder}} The GCP corpus was available to us in a pre-tokenized and lowercased form.

We implemented our method on top of an open-source implementation of the Transformer model \citep{NIPS2017_7181} in the version 1.6 branch of \textit{tensor2tensor}.\footnote{\url{https://github.com/tensorflow/tensor2tensor}}  For training, we used the default model setting corresponding to \textit{transformer\_base\_single\_gpu} in the implementation, except that we varied the number of sub-words and training iterations depending on the dataset.
We used the tensor2tensor internal sub-word segmenter for simplicity. The settings of sub-word vocabularies and training iterations are given in \Tab{datasets}. Note that for the GCP task, we chose the vocabulary size used by \citet{W18-2707}, and trained \unidir{En}{Ja} models for 60k iterations whereas we trained \unidir{Ja}{En} models for 120k iterations.

We averaged the last 10 check-points and decoded the test sentences with a beam size of 4 and length penalty of $\alpha=0.6$ for the KFTT \unidir{Ja}{En} task and $\alpha=1.0$ for the rest. We evaluated our models using the BLEU metric \citep{Papineni:2002:BMA:1073083.1073135} implemented in tensor2tensor as \textit{t2t\_bleu}: case-insensitive and tokenized BLEU.

\subsection{Reduction in Model Size}
\label{section:model_size}

\Tab{model-parameters} compares the number of parameters in the vanilla and RS models in all the translation tasks.  A 1-layer RS model is identical to a 1-layer vanilla model.
For the same dataset, the number of parameters for the two translation directions can be different due to the implementation of the sub-word mechanism in tensor2tensor, which does not produce vocabularies of exactly the same sizes as we specified (\Tab{datasets}).

Whereas the numbers of parameters of vanilla NMT models increase according to the increase of the number of layers, those for RS models do not, leading to a significant difference.
We observed higher percentages of parameter reduction for the settings with smaller vocabularies; compared to the 6-layer vanilla NMT, approximately 65\%, 55\%, and 40\% of the parameters were reduced for the KFTT, GCP, and ASPEC tasks, respectively.  This is because the bulk of the parameters in NMT are used for mapping between sub-words and hidden representation at the embedding and softmax layers.

\begin{figure*}[t]
    \centering
    \begin{subfigure}[t]{0.48\textwidth}
    \includegraphics*[width=\textwidth]{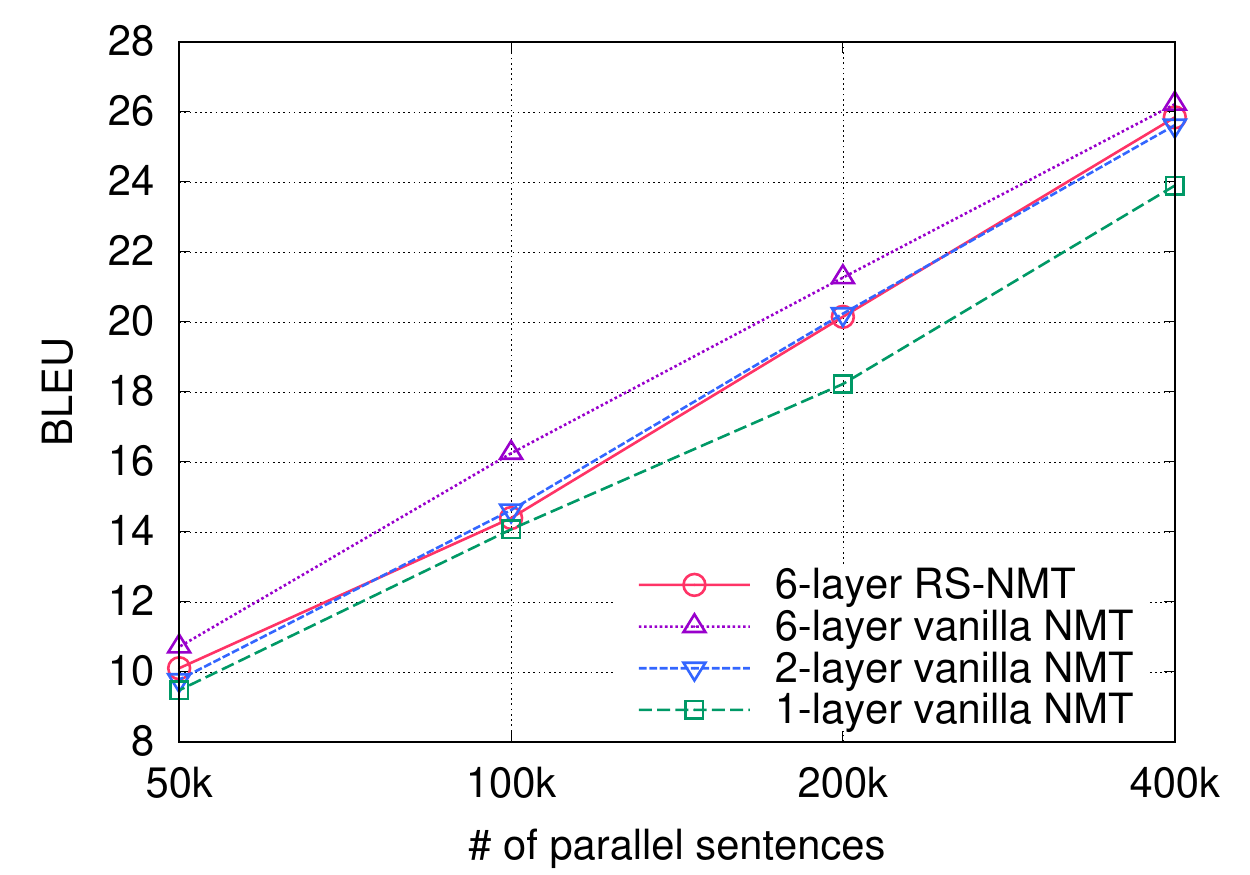}
    \caption{GCP \unidir{En}{Ja} task.}
    \end{subfigure}
    \begin{subfigure}[t]{0.48\textwidth}
    \includegraphics*[width=\textwidth]{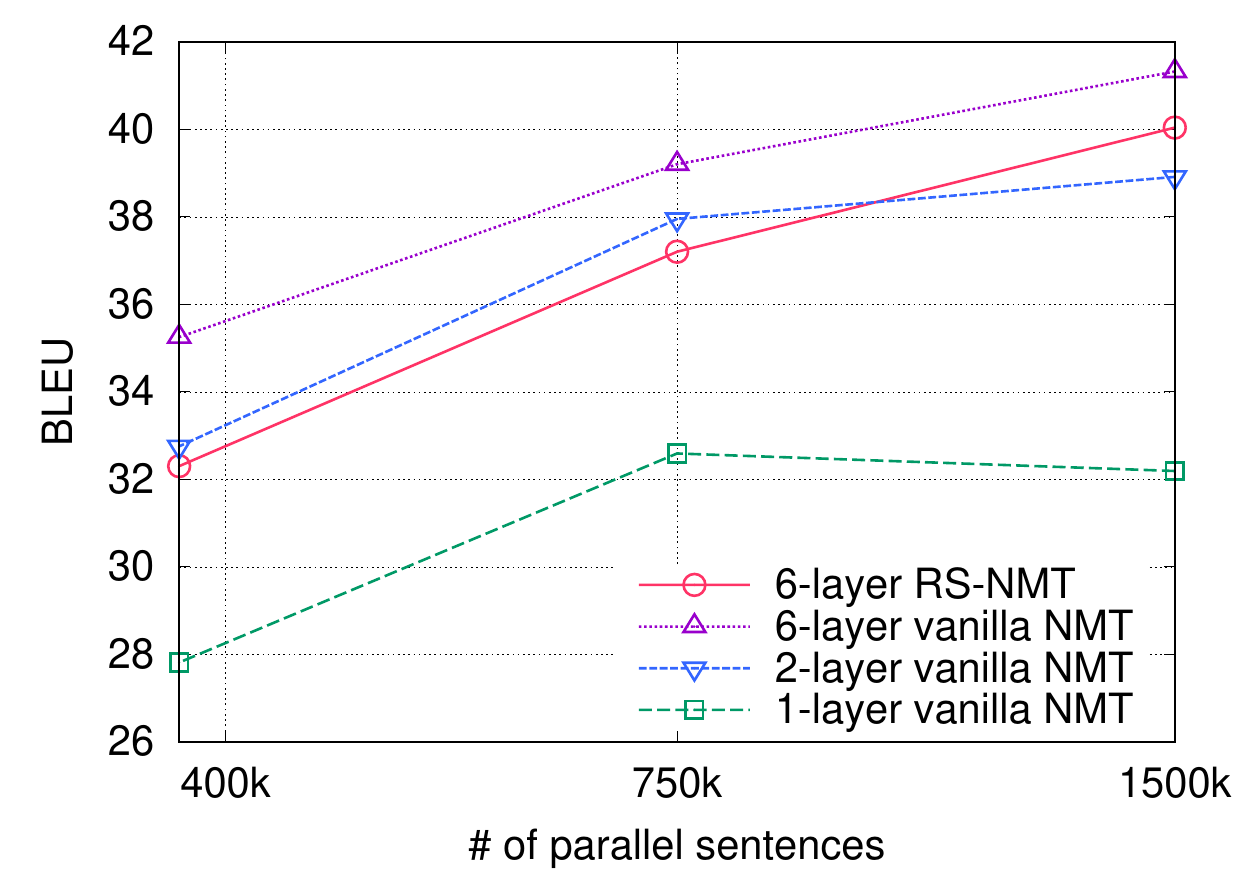}
    \caption{ASPEC \unidir{En}{Ja} task.}
    \end{subfigure}
    \caption{BLEU scores of the 6-layer models trained on different sizes of data.}
    \label{figure:corpus-reduction}
\end{figure*}

\subsection{Translation Quality}
\label{section:quality}

\Tab{direct-results} summarizes the translation quality in BLEU for all the datasets.
In what follows, we first compare the vanilla and RS models, and then describe three additional experiments: learning-curve experiment, training with back-translations, and an exploration of much deeper model.

\subsubsection{Recurrently Stacked Models vs. Vanilla Models}
\label{section:main_results}

In general, no matter which dataset was used, the translation quality improved as the same parameters were recurrently used in a depth-wise fashion. In all the translation tasks, except ASPEC \unidir{Ja}{En}, the 6-layer RS-NMT model was better than the corresponding 2-layer vanilla NMT model that has more parameters, meaning that our RS-NMT models can compensate for the lack of parameters. However, even though the performance of a 6-layer RS-NMT model approaches that of the 6-layer vanilla NMT model, there are still significant differences in the two KFTT tasks and the two ASPEC tasks.
For the resource-richest ASPEC task, the 6-layer RS-NMT models were better than the 1-layer vanilla NMT models by 3.92 and 7.85 BLEU points, respectively, and they were worse than the 6-layer vanilla NMT models by only 1.57 and 1.28 BLEU points, respectively. Similar observations hold for the GCP and KFTT tasks. Even though a drop in translation quality is undesirable, in our opinion, these drops are not catastrophic given the reduction in the number of parameters.

Most existing studies assume that the improvement in performance of deep stacked models comes from the large number of parameters and hence the complex representations learned. However, even though an RS-NMT model has exactly the same number of parameters as a 1-layer vanilla NMT model, its representational power seems to be much higher. It is clear that the continuous representations undergo a number of non-linear transformations (linear transformations and non-linear activations) thanks to the RS of layers.  The same happens in vanilla NMT models except that more parameters are involved per transformation at each layer. Thus, our comparison suggests that it is the non-linear transformations that are responsible for the bulk of the improvement in deep stacked models. In our opinion, this is an important observation in deep learning research and deserves further exploration.

\subsubsection{Effect of Corpora Sizes on Translation Quality}
\label{section:corpus_size}

\Fig{corpus-reduction} shows the difference in the performance between the RS-NMT and the vanilla NMT when we vary the size of the training data for the same datasets. We plot the BLEU scores for the GCP \unidir{En}{Ja} and ASPEC \unidir{En}{Ja} tasks. Whereas it is obvious that reducing the size of training data deteriorates the BLEU scores, the trends in terms of difference in performance between the RS-NMT and vanilla NMT remain almost the same: i.e., 6-layer RS model is much better than 1-layer vanilla model, even though it is quite clear that the reduction in the number of parameters can come with a drop in translation quality when compared to 6-layer vanilla model.
As such, we can safely say that our proposed method is independent of corpus size and domain.

\subsubsection{Training with Back-Translation}

Since back-translation is one of the most reliable ways to boost translation quality \cite{sennrich-haddow-birch:2016:P16-11}, we evaluated its impact on our proposed method.
We experimented with the GCP tasks, since 1.55M sentences in the Japanese and English monolingual corpora for the same domain \cite{W18-2707} are available as the source to generate back-translations.
First, we generated pseudo-parallel data by back-translating the monolingual sentences using the 1-layer models for the opposite translation direction.\footnote{For example, we used the 1-layer \unidir{Ja}{En} model in order to translate the Japanese monolingual sentences for training new \unidir{En}{Ja} models. Even though our original intention to use the 1-layer model is its fast decoding, as the 1-layer RS-NMT model is identical to the 1-layer vanilla NMT model, this eventually makes a fair comparison: all the models for the same translation direction are trained on exactly the same pseudo-parallel data.} We then trained, from scratch, the 6-layer vanilla NMT and up to 6-layer RS-NMT models on the mixture of the pseudo-parallel and the original parallel data. To compensate for the additional data, we trained both the \unidir{Ja}{En} and the \unidir{En}{Ja} models for 200k iterations (cf. \Tab{datasets}).

\Tab{backtransresults} provides the results. Despite no increase in the number of parameters, the presence of back-translated data improved the translation quality for all the translation tasks. The 2-layer RS-NMT trained using additional back-translated data already outperformed the 6-layer vanilla NMT models trained only on the original parallel corpus. Note that we used 1-layer models to generate the back-translated data and it is natural to expect that using deeper models will give better back-translated data leading to further improvements in translation quality.

\begin{table}[t]
\centering
\small
\begin{tabular}{c|c|c|c|c}
\hline
\multirow{2}{*}{Model}
& \multicolumn{2}{c|}{\unidir{Ja}{En}}
& \multicolumn{2}{c}{\unidir{En}{Ja}}\\\cline{2-5}
& No & Yes & No & Yes \\\hline

\multicolumn{5}{l}{\textbf{(a) Vanilla NMT}}\\\hline
6 & 24.67 & \textbf{25.91} & 26.22 & \textbf{28.75} \\\hline

\multicolumn{5}{l}{\textbf{(b) RS-NMT}}\\\hline
1 & 21.95 & \textbf{23.90} & 23.89 & \textbf{25.47} \\
2 & 23.24 & \textbf{24.79} & 24.47 & \textbf{26.56} \\
3 & 23.42 & \textbf{24.79} & 25.02 & \textbf{26.66} \\
4 & 24.33 & \textbf{25.17} & 25.28 & \textbf{27.31} \\
5 & 23.95 & \textbf{24.92} & 25.38 & \textbf{27.08} \\
6 & 24.36 & \textbf{25.82} & 25.84 & \textbf{27.55} \\
\hline
\end{tabular}
\caption{BLEU scores obtained using back-translated data for the GCP tasks. The ``Yes'' and ``No'' columns indicate the involvement of back-translated data. The highest BLEU score in each configuration is in bold.}
\label{table:backtransresults}
\end{table}

\subsubsection{Limits of Recurrent Stacking}
\label{section:limits}

In \Sec{main_results}, we observed that non-linear transformations cover the bulk of improvement when testing with up to 6-layer models.  However, similarly to vanilla models, we expect that the increase in the number of recurrence does not always bring improvement, i.e., there must be a limit.
To determine if there is a depth-benefit trade-off with RS-NMT modeling, we trained a 24-layer RS-NMT model for the GCP \unidir{En}{Ja} task. We obtained a BLEU score of 26.04, with a non-significant 0.20 point improvement over the 6-layer RS-NMT model which has a BLEU score of 25.84. One possible reason for this is that the gradients vanish during back-propagation due to the extreme depth. We leave the application of sophisticated training techniques for the future. Note that a very deep RS-NMT model with high translation quality is more valuable than its vanilla counterpart, because the size of an RS-NMT model always remains the same whereas the size of the vanilla NMT model continues to grow according to the increase of the number of layers. Work on identifying the equilibrium of RS models \cite{DBLP:conf/nips/BaiKK19} has shown that it is possible to predict the outputs of extremely deep RS layers. This means that it should be possible to simulate extremely deep RS-NMT models and their improved performance without being slowed down by their computational complexity.

\section{Training Recurrent Stacking Models through Transfer Learning}
\label{section:transfer_learning}

Our empirical experiments presented in \Sec{experiments} revealed that the RS models, despite having the same computational complexity\footnote{Both vanilla and RS models perform the same number of matrix multiplications.} as vanilla models, tend to suffer from reduced performance, due to a significant reduction of the number of parameters.
To obtain compact models with a simple architecture but with comparable performance to vanilla models, this section examines ways to train RS models through transfer learning from a pre-trained vanilla model instead of merely relying on the parallel data as in \Sec{experiments}. 

\begin{figure*}[t]
    \centering
    \begin{subfigure}[b]{0.32\textwidth}\centering
    \includegraphics*[scale=.45]{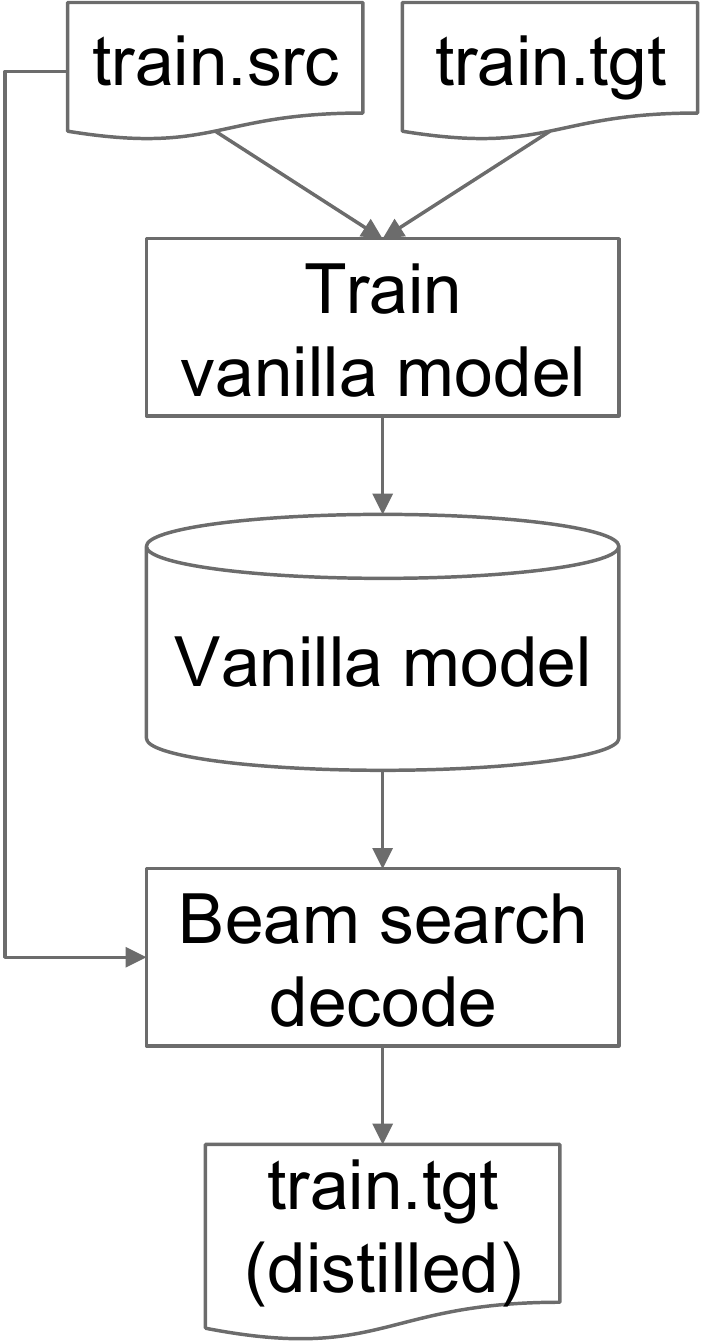}
    \caption{Preparation}
    \label{figure:preparation}
    \end{subfigure}
    \begin{subfigure}[b]{0.32\textwidth}\centering
    \includegraphics*[scale=.45]{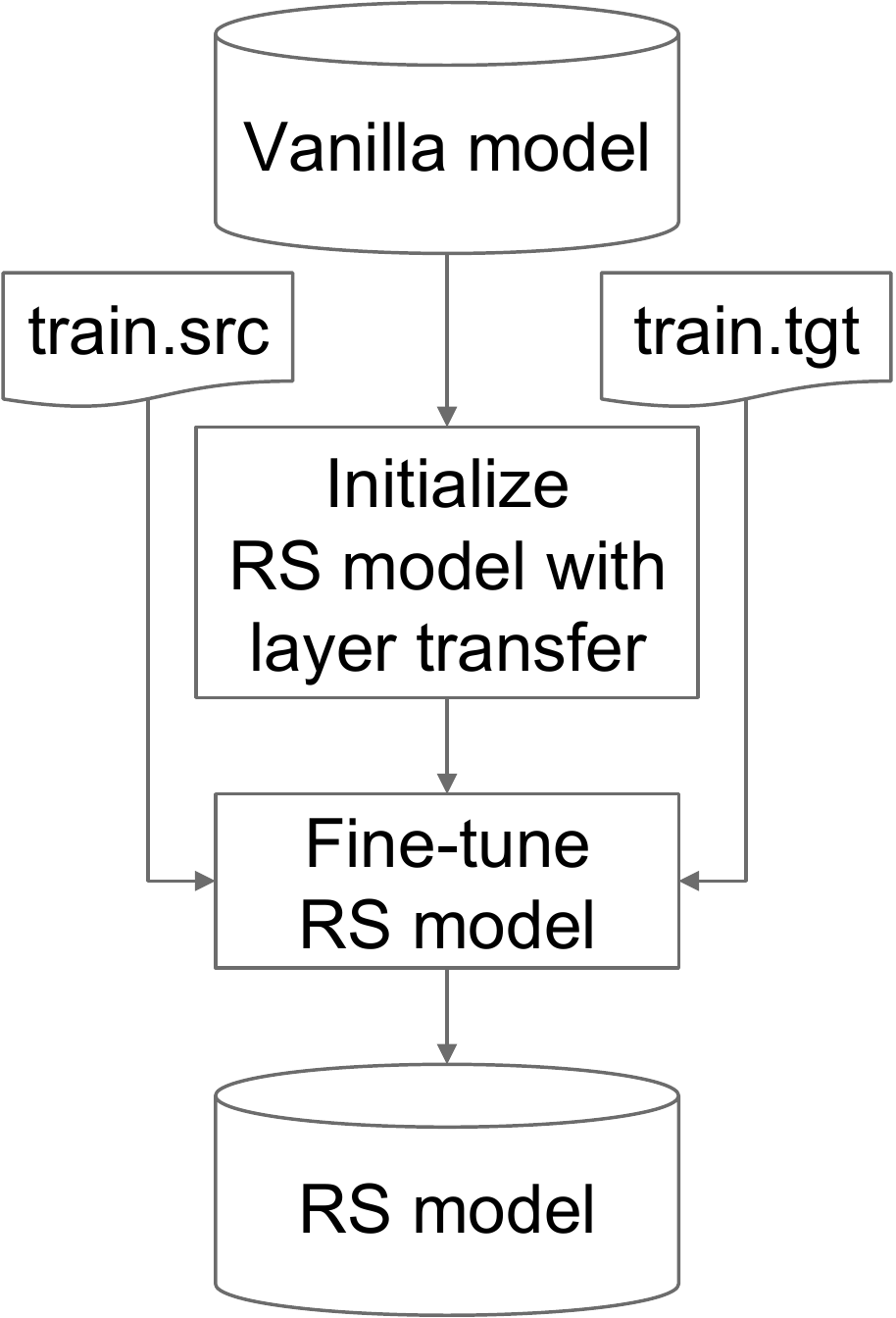}
    \caption{Layer transfer}
    \label{figure:transfer}
    \end{subfigure}
    \begin{subfigure}[b]{0.32\textwidth}\centering
    \includegraphics*[scale=.45]{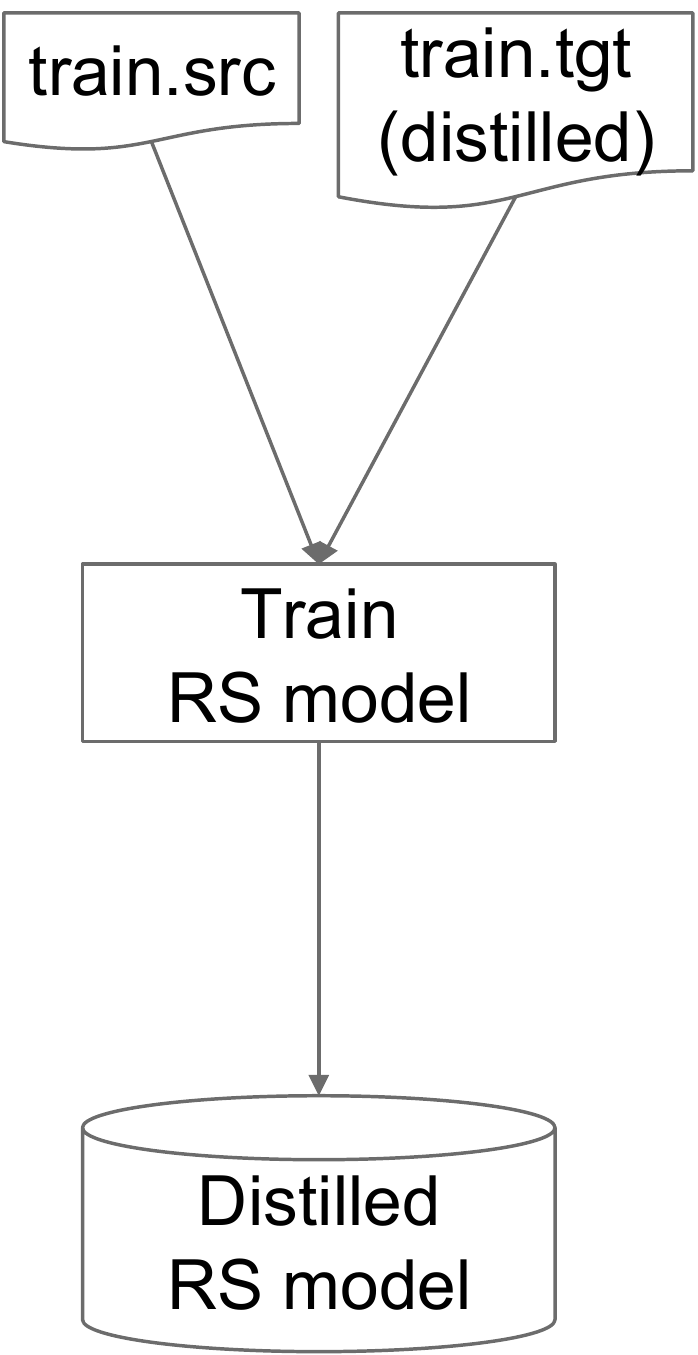}
    \caption{Sequence distillation}
    \label{figure:distillation}
    \end{subfigure}
    \caption{Overview of techniques used for bridging the gap between RS and vanilla models. (a)~Training a ``vanilla model'' and using it to generate ``train.out (distilled)'' for the inputs of the training data. (b)~Training an RS model using the pre-trained parameters of the ``vanilla model'' for initialization. (c)~Training an RS model from scratch on the pair of the inputs of the training data and decoded outputs, i.e., ``train.out (distilled).''}
    \label{figure:overview-methodologies}
\end{figure*}

\subsection{Transfer Learning Methods}

Transfer learning for neural network models involves utilizing strong pre-trained teacher model(s) to transfer its capabilities to a student model that is either small or has less training data. The underlying principle is to leverage the priors from the pre-trained model(s). In this paper, we explore two ways of transferring knowledge: a simple layer transfer followed by fine-tuning and sequence-level knowledge distillation. The former is an explicit way of transferring knowledge whereas the latter is an implicit way of transferring knowledge. We also explore the effects of combining both methods.
\Fig{overview-methodologies} illustrates an overview of the procedure.

\subsubsection{Layer transfer and fine-tuning}
\citet{DBLP:conf/emnlp/ZophYMK16:original} have proposed fine-tuning as a way to improve NMT in a low-resource scenario, such as Hausa-to-English translation. First, a teacher model is trained on a high-resource language pair like French-to-English. The parameters of this model are then used to initialize those of a student Hausa-to-English model, and the training is resumed on the Hausa--English parallel corpus.
In our case, as depicted in \Fig{transfer}, the teacher and student models are vanilla and RS models, respectively, but the parallel corpus is the same. Whereas the embeddings and softmax layers of the RS model can be initialized with those of the vanilla model, it has a single encoder and decoder layer unlike the vanilla model which has six. It is unclear as to which particular layer in the vanilla model is most useful a priori and we can exploit only one layer, each from the encoder and decoder, of the vanilla model to initialize the corresponding encoder and decoder layers of the RS model. As such, we explore all possible models with different choices of layers in the encoder and decoder of the vanilla model. After initialization, we continue training on the parallel corpus till convergence.

\subsubsection{Knowledge Distillation}

Knowledge distillation \citep{DBLP:journals/corr/HintonVD15} uses the probability distribution generated by a teacher model
as a label for training a student model.
The rationale is that a probability distribution as a smoothed label enables a student model to learn as well as a teacher model that has already been trained using one-hot labels. However, this gives minimal benefits for sequences, because sequence models traditionally use teacher-forcing\footnote{For clarity, teacher-forcing is an input-feeding technique for training or decoding neural network models, whereas teacher models are models whose behavior is to be learned by student models. As such, teacher-forcing can be used to train models regardless of whether they are teacher or student.} during training \cite{journals/corr/BengioVJS15}. Teacher-forcing implies the use of the gold label instead of the predicted label as an input to the model in order to predict the next label. As such, using teacher-forcing prevents the student model from learning decoding-time behavior of the teacher model and the result is distillation being done at the word level. This also involves loading two models into memory and several implementation changes.

The problems caused by teacher-forcing can be mitigated by greedy or beam search decoding of the teacher model while training the student model.  However, this can slow down training substantially. Note that once the teacher model has been trained, its decoding behavior will remain the same, and hence it is faster to generate distilled sequences before training the student model.
In our context, we regard vanilla and RS models as teacher and student models, respectively, as depicted in Figures~\ref{figure:preparation} and \ref{figure:distillation}.
Note that sequence-level distillation is a forward translation technique, which has already been used for self-learning of machine translation models \citep{ueffing:07}.

Combining word- and sequence-level distillation methods \cite{kim-rush-2016-sequence} has been shown to improve the performance of smaller models, even though the bulk of the improvement comes from sequence-level distillation. We leave the combination for our future work.

\subsection{Experiments}
\label{section:transfer_results}

We conducted experiments to confirm the efficacy of the two transfer learning methods. We first compare the impact of these methods on an experiment for all translation tasks (\Sec{impact}). We then determine whether they are complementary, taking an example task (\Sec{complementarity}).
This is followed by focused analyses of our approaches on different sets of tasks, as applicable, where we
\begin{itemize}
\item determine when and where sequence distillation may or may not be useful (\Sec{limits-sd}),
\item discuss the impact of sequence distillation on decoding efficiency (\Sec{speedup}), and
\item perform a cost-benefit analysis of translation quality, model size, and decoding speed to determine when and where RS models should be preferred over vanilla models (\Sec{costbenefitanalysis}).
\end{itemize}

\subsubsection{Impact of Transfer Learning on Translation Quality}
\label{section:impact}

To confirm that the transfer learning methods can help us train compact RS models with performance comparable to vanilla models, we trained and evaluated the following three new types\footnote{Since our aim is to obtain a compact model with the RS approach, we do not evaluate the configurations where RS and vanilla models are regarded as teacher and student models, respectively.} of NMT models for all translation tasks with the same settings as those in \Sec{experiments}.
\begin{description}
    \item[(c)~RS-NMT fine-tuned:] 6-layer\footnote{Note that we avoided training models with fewer layers (recurrences) for this specific scenario as the number of results become too much to present.} RS-NMT models where the encoder and decoder layers were initialized with $l$-th ($l\in\{1,2,6\}$)\footnote{Note that we do not assume the same depth of the encoder and decoders layers in the vanilla model must be used to initialize those in the RS model.} encoder and decoder layers of the pre-trained 6-layer vanilla NMT model, respectively.
    \item[(d)~RS-NMT distilled:] $k$-layer ($k\in\{1,2,6\}$) RS-NMT models trained on the pseudo-parallel data generated by the pre-trained 6-layer vanilla NMT model.
    \item[(e)~Vanilla-NMT distilled:] $k$-layer ($k\in\{1,2,6\}$) vanilla NMT models trained on the pseudo-parallel data generated by the pre-trained 6-layer vanilla NMT model.
\end{description}

\begin{table*}[t]
\centering
\small
\begin{tabular}{c|c|c|c|c|c|c}
\hline
\multirow{2}{*}{Model} & \multicolumn{2}{c|}{GCP} & \multicolumn{2}{c|}{KFTT} & \multicolumn{2}{c}{ASPEC} \\ \cline{2-7} 
 & \unidir{Ja}{En} & \unidir{En}{Ja} & \unidir{Ja}{En} & \unidir{En}{Ja} & \unidir{Ja}{En} & \unidir{En}{Ja} \\ \hline

\multicolumn{7}{l}{\textbf{(a) Vanilla NMT}}\\\hline
1 & 21.95 & 23.89 & 21.64 & 25.00 & 23.28 & 32.19 \\
2 & 24.23 & 25.62 & 24.14 & 30.05 & 28.06 & 38.91 \\
6 & 24.67 & 26.22 & 27.19\sig & \textbf{32.72}{\sig} & 28.77{\sig} & \textbf{41.32}{\sig} \\ \hline

\multicolumn{7}{l}{\textbf{(b) RS-NMT scratch}}\\\hline
1 & 21.95 & 23.89 & 21.64 & 25.00 & 23.28 & 32.19 \\
2 & 23.24 & 24.47 & 24.50 & 28.53 & 27.84 & 38.54 \\
6 & 24.36 & 25.84 & 26.51 & 30.83 & 27.20 & 40.04 \\ \hline

\multicolumn{7}{l}{\textbf{(c) RS-NMT fine-tuned (only 6-layer models)}}\\\hline
1 & 23.86 & 25.33 & 25.89 & 30.93 & 27.44 & 39.16 \\
2 & 24.10 & 25.70 & 26.63 & 31.68{\sig} & 27.52 & 40.07 \\
6 & 22.95 & 24.63 & 26.09 & 30.58 & 28.69{\sig} & 39.75 \\ \hline

\multicolumn{7}{l}{\textbf{(d) RS-NMT distilled}}\\\hline
1 & 23.25 & 24.20 & 23.23 & 26.47 & 27.81 & 35.71 \\ 
2 & 24.52 & 25.78 & 25.39 & 30.51 & 29.62{\sig} & 38.86 \\ 
6 & 25.20{\sig} & 26.94{\sig} & 27.11{\sig} & 31.52{\sig} & 29.18{\sig} & 40.43 \\ \hline

\multicolumn{7}{l}{\textbf{(e) Vanilla NMT distilled}}\\\hline
1 & 23.25 & 24.20 & 23.23 & 26.47 & 27.81 & 35.71 \\ 
2 & 24.91 & 26.38{\sig} & 26.61 & 30.26 & 29.23\sig & 40.11 \\ 
6 & \textbf{25.75}{\sig} & \textbf{27.44}{\sig} & \textbf{27.53}{\sig} & 32.58{\sig} & \textbf{29.68}{\sig} & 40.92{\sig} \\ \hline
\end{tabular}
\caption{BLEU scores obtained in our experiments. The definition of numbers in the leftmost column are as follows: (a) and (e)~the number of different encoder and decoder layers, (b) and (d)~the number of recurrence, and (c)~the depth of encoder and decoder layers of the pre-trained 6-layer vanilla NMT model used for initializing the RS-NMT model. Scores in bold are the ones with highest BLEU score for each translation direction. Scores marked with ``\sig'' are the ones that are statistically significantly (bootstrap re-sampling with $p<0.05$) better than all the RS-NMT models trained from scratch i.e., (b).}
\label{table:all-results}
\end{table*}

\Tab{all-results} shows the results of these models along with some excerpts for models (a) and (b) from \Tab{direct-results}.  Unless mentioned otherwise, we compare models that use 6-layer vanilla or RS layers.
It is clear that transfer learning helps improve the performance of an RS-NMT model compared to when it is trained from scratch, i.e., models (b). Models (d) trained via sequence distillation gave better results than models (c) trained via fine-tuning for all the tasks, except for the KFTT \unidir{En}{Ja} task. Furthermore, models (d) outperformed the vanilla NMT models for the two GCP tasks and the ASPEC \unidir{Ja}{En} task.
We also trained vanilla NMT models on the distillation data, i.e., models (e), for comparison, and noticed that they sometimes improve over vanilla NMT models trained on the original data, i.e., models (a). They are slightly to significantly better (approximately 1.0 BLEU in some tasks) than their corresponding distilled RS-NMT models, i.e., models (d), owing to the much large number of parameters (\Tab{model-parameters}). We will discuss this in further detail in \Sec{costbenefitanalysis}.

\subsubsection{Complementarity of Two Methods}
\label{section:complementarity}

To determine whether the two transfer learning methods are complementary, we trained and evaluated another type of the RS-NMT models.
\begin{description}
\item[(f)~RSNMT-NMT fine-tuned+distilled:] $k$-layer ($k\in\{1,2,3,4,5,6\}$) RS-NMT models trained on the pseudo-parallel data generated by the pre-trained 6-layer vanilla NMT model, with the encoder and decoder RS layers initialized with the $l$-th ($l\in\{1,2,3,4,5,6\}$) encoder and decoder layers of the same vanilla NMT model, respectively.
\end{description}

\begin{table*}[t]
\centering
\small
\begin{tabular}{c|c|c|c|c|c|c|c}
\hline
$l{\backslash}k$ & 1 &2 &3 &4 &5 &6 &(c) \\ \hline
1 & 24.60 & 25.68 & 25.62 & 26.75 & 26.15 & 26.20 & 25.33 \\
2 & 24.70 & 26.33 & 25.64 & 26.57 & 26.53 & 26.23 & 25.70 \\
3 & 24.70 & 26.02 & 26.48 & 26.33 & 26.68 & 26.67 & 25.66 \\
4 & 24.51 & 26.06 & 26.21 & 26.38 & 26.45 & 26.49 & 26.22 \\
5 & 24.64 & 26.20 & 25.98 & 25.93 & 25.89 & 26.24 & 24.97 \\
6 & 24.57 & 26.03 & 26.28 & 25.86 & 26.51 & 26.16 & 24.63 \\
\hline
(d) & 24.20 & 25.78 & 25.97 & 26.27 & 26.00 & 26.94 \\
\cline{1-7}
\end{tabular}
\caption{BLEU scores for the GCP \unidir{En}{Ja} task achieved by combining the two transfer learning methods, i.e., fine-tuning and sequence distillation, for training RS-NMT models. $k$ denotes the number of recurrence of the RS-NMT model and $l$ the depth of encoder and decoder layers of the pre-trained 6-layer vanilla NMT model used for initializing the RS-NMT model. The the rightmost column and last row respectively show for reference the results of model (c), i.e., 6-layer RS-NMT models trained with parameter initialization using one of 6 layers but without sequence distillation and model (d), i.e., 1- to 6-layer RS-NMT models trained via sequence distillation without any parameter initialization.}
\label{table:mixed-ft-distillation-results}
\end{table*}

\Tab{mixed-ft-distillation-results} gives the results for the GCP \unidir{En}{Ja} task, which we take as an example for completion.  When the number of recurrence ($k$) is small, initializing the RS layer yields slight performance gains over the corresponding model (d) without initialization; however, 6-layer RS models no longer benefit from parameter initialization. Over the models (c) with the same number of recurrence ($k=6$), sequence distillation brings consistent improvements. 

Consider the case of $k=2$ and $l=2$: a 2-layer RS-NMT model initialized with the 2nd encoder/decoder layers and fine-tuned on the data for sequence distillation.
This model achieves a BLEU of 26.33 which is better than its non-initialized counterpart, i.e., model (d), with a BLEU of 25.78. This shows that using the combination of sequence distillation and fine-tuning can enable one to train shallower RS models with performance that is significantly better than deeper RS models that do not use any distillation or transfer learning.

Even though we do not report on the results for all translation tasks due to the sheer number of combinations, the results so far confirm that combination of these two transfer learning methods is beneficial and they appear to complement each other.

\subsubsection{Limits of Sequence Distillation}
\label{section:limits-sd}

To investigate the reasons why sequence distillation is not always better than fine-tuning, we measured using BLEU the similarity between (i)~translations of the training data generated as the pseudo-target sentences used for sequence distillation and (ii)~reference translations in the original training data. \Tab{distillation} shows the BLEU scores of a random sample of 10,000 sentences from the training data that we translated for sequence distillation. In the ASPEC \unidir{En}{Ja} task, the BLEU score was almost 60. For reference, the BLEU scores for all other settings were significantly lower. The higher the BLEU score, the higher the similarity of the translated sentences with the references. One of the reasons why sequence distillation can enable smaller models to perform better, according to \citet{kim-rush-2016-sequence}, is because the translated sentences are supposed to be simplified and hence inaccurate versions of the references. As it is easier for smaller models to learn simpler data, the increased similarity between the translations and references prevents from performing better. Consequently, it seems logical that sequence distillation would not help the performance of RS-NMT models in this particular setting. Despite ASPEC being a high-resource setting, the vanilla NMT model seems to learn from the training data much better than for the other datasets. We suppose that stronger regularization during training can mitigate this, and will explore this and its impact on sequence distillation in the future.

\begin{table}[t]
\centering
\small
\begin{tabular}{c|c|c}
\hline
Dataset & \unidir{Ja}{En} & \unidir{En}{Ja}\\\hline
GCP   & 45.54 & 51.70 \\
KFTT  & 37.91 & 43.42 \\
ASPEC & 51.14 & 59.70 \\
\hline
\end{tabular}
\caption{BLEU scores for 10,000 random samples in the training data, as the similarity between original parallel data and the pseudo parallel data used for sequence distillation.}
\label{table:distillation}
\end{table}

Another observation is that, due to distillation, the performance does not drastically improve by using two or more layers. This applies to both vanilla and RS NMT models. We also confirmed one of the observations of \citet{kim-rush-2016-sequence}: the distilled models can give better performance than their non-distilled counterparts. Distillation benefits both vanilla and RS NMT models and distilled vanilla NMT models are the best, although distilled RS-NMT models are better than or as good as vanilla NMT models trained from scratch. In the end, the gap in translation quality between vanilla and RS NMT models is approximately 1.0 BLEU.  However, in the big picture, we do not think that this is a serious matter. Given that RS-NMT models are more compact than vanilla NMT models, they are more attractive in our opinion.

\subsubsection{Faster Decoding Through Transfer Learning}
\label{section:speedup}

The RS-NMT models trained via transfer learning are typically more around 1.0 to 2.0 BLEU points better than their counterparts trained without transfer learning, irrespective of the number of RS layers.  This enables us to save decoding time by the following two approaches.

\paragraph{Use of greedy decoding:}

Note that all BLEU scores so far presented are for translations derived via beam search with a beam size of 4.  \Tab{beam-vs-greedy} compares BLEU scores and decoding times of two sets of RS-NMT models, taking the GCP \unidir{En}{Ja} as an example task.\footnote{For the sake of presentation, we do not give the results for all tasks, but \Tab{all-results} suggests that the trends for GCP \unidir{En}{Ja} task are applicable for most settings.} One is RS-NMT models trained on the original data, i.e., models (b), used for beam search decoding, and the other is RS-NMT models trained via sequence distillation, i.e., models (d), used for beam search decoding and greedy decoding. Whereas there is no significant difference between the first and the last configurations regarding BLEU scores, beam search takes twice the time as greedy search. This means that we can save a substantial amount of time by using greedy search thanks to sequence distillation.

\paragraph{Use of shallower models:}

According to \Tab{beam-vs-greedy}, for the GCP \unidir{En}{Ja} task, beam search with 6-layer RS-NMT model trained on the original data, i.e., model (b), takes 83.29s to generate translations with BLEU of 25.84.  Thanks to sequence distillation, a shallower 2-layer RS-NMT model (d) is now able to generate translations with the comparable quality, BLEU of 25.78, spending only 39.50s with the same width of beam, which is less than half of time of the 6-layer models.

\begin{table*}[t]
  \centering
  \small
  \begin{tabular}{l|cc|cc|cc}
    \hline
    \multirow{3}{*}{$k$} &\multicolumn{2}{c|}{Model (b)} &\multicolumn{2}{c|}{Model (d)} &\multicolumn{2}{c}{Model (d)}\\
    &\multicolumn{2}{c|}{w/ beam search} &\multicolumn{2}{c|}{w/ beam search} &\multicolumn{2}{c}{w/ greedy search}\\\cline{2-7}
    &BLEU &Time (s) &BLEU &Time (s) &BLEU &Time (s)\\
    \hline
    1 &23.89 &36.96 &24.20 &36.96 &23.66 &18.41\\
    2 &24.47 &39.50 &25.78 &39.50 &24.52 &23.78\\
    6 &25.84 &83.29 &26.94 &83.29 &25.57 &46.22\\
    \hline
  \end{tabular}
  \caption{Comparison of (b) RS-NMT models directly trained and (d) RS-NMT models trained via sequence distillation used with different search methods for the GCP \unidir{En}{Ja} task.  Decoding time includes model loading time and writing time. $k$ denotes the number of recurrence.}
  \label{table:beam-vs-greedy}
\end{table*}

\subsubsection{Vanilla vs. RS: Cost Benefit Analysis}
\label{section:costbenefitanalysis}

We now present a cost benefit analysis of vanilla and RS models in various model size settings. We already know that in the absence of transfer learning methods, such as sequence distillation or fine-tuning, vanilla models always outperform RS models. We thus focus on comparing these two types of models when using sequence distillation as a transfer learning method.\footnote{We chose sequence distillation and not fine-tuning, mainly because it is straightforward to apply sequence distillation than fine-tuning since the latter involves determining optimal vanilla model layers prior to fine-tuning.} For this analysis, we chose the ASPEC \unidir{Ja}{En} task, because it is one of the cases where the RS model performs on par\footnote{Note that in the results so far, the relative gap between RS and vanilla models is significantly smaller in high-resource settings, such as ASPEC.} with the vanilla model as in \Tab{all-results}. Our objective is to identify precise conditions where RS should be preferred over vanilla models.
We trained (whenever possible) vanilla and RS models using 1 to 6 encoder-decoder layers with the dimensionality of hidden layers of 64, 128, 258, 512, 1,024, and 2,048.  In all these configurations, the filter size was consistently set to four times of the hidden layer, as the Transformer Base model.

\begin{table*}[t]
\centering
\small
\begin{tabular}{c|c|c|c|c|c|c}
\hline
Hidden-Filter & 64-256 & 128-512 & 256-1,024 & 512-2,048 & 1,024-4,096 & 2,048-8,192 \\
\hline\hline

Layer & \multicolumn{6}{c}{Model Sizes}\\\hline
1, RS
  & 6.4M & 13.0M & 27.0M & 57.7M & 130.1M & 319.0M \\\hline
2 & 6.5M & 13.5M & 28.8M & 65.0M & 159.5M & n/a    \\\hline
3 & 6.6M & 14.0M & 30.7M & 72.4M & 188.9M & n/a    \\\hline
4 & 6.7M & 14.4M & 32.5M & 79.7M & 218.3M & n/a    \\\hline
5 & 6.8M & 14.9M & 34.4M & 87.1M & 247.7M & n/a    \\\hline
6 & 7.0M & 15.3M & 36.2M & 94.4M & n/a    & n/a    \\\hline
\hline

Layer & \multicolumn{6}{c}{BLEU Scores}\\\hline
1
& 14.61/16.04 & 20.61/21.95 & 24.81/25.94 & 26.99/27.81 & 27.68/28.31 & 27.74/28.17 \\\hline

\multirow{2}{*}{2}
& 16.50/18.53 & 24.70/25.52 & 27.86/28.46 & \bf 29.11/29.62 & \bf 29.39/29.74 & \multirow{2}{*}{n/a} \\
& 20.54/21.23 & 26.49/27.20 & 28.58/28.92 & 28.88/29.23 & 29.12/29.47 \\\hline

\multirow{2}{*}{3}
& 17.80/18.67 & 25.80/26.49 & 28.18/28.83 & \bf 29.14/29.57 & 29.24/29.54 & \multirow{2}{*}{n/a} \\
& 22.00/23.12 & 27.34/27.91 & 29.31/29.36 & 28.95/29.45 & 29.56/29.88 \\\hline

\multirow{2}{*}{4}
& 19.86/21.26 & 26.54/27.21 & 28.35/28.74 & 28.97/29.39 & 28.90/29.28 & \multirow{2}{*}{n/a} \\
& 24.27/25.09 & 27.73/28.02 & 29.21/29.48 & 29.22/29.70 & 29.25/29.65 \\\hline

\multirow{2}{*}{5}
& 20.68/21.98 & 26.74/27.56 & 28.53/29.00 & 28.68/29.28 & \bf 29.56/30.14 & \multirow{2}{*}{n/a} \\
& 24.22/25.29 & 28.26/28.65 & 29.33/29.62 & 29.17/29.68 & 29.50/29.97 \\\hline

\multirow{2}{*}{6}
& 21.20/22.69 & 27.06/27.52 & 28.68/28.96 & 28.70/29.18 & 29.05/29.69 & \multirow{2}{*}{n/a} \\
& 25.47/26.13 & 27.91/28.55 & 28.98/29.40 & 29.27/29.68 & n/a \\\hline
\hline

Layer
& \multicolumn{6}{c}{Decoding Times}\\\hline

1
& 37.4/38.9 & 32.5/38.5 & 31.2/39.8 & 32.4/47.2 & 37.1/74.7 & 59.4/162.8 \\\hline

\multirow{2}{*}{2}
& 41.1/55.1 & {\bf 36.2}/48.4 & {\bf 36.9}/53.0 & 41.3/56.8 & \bf 46.9/94.6 & \multirow{2}{*}{n/a} \\
& 38.0/48.9 & 36.7/46.8 & 38.1/51.7 & 41.2/55.1 & 51.0/111.3 \\\hline

\multirow{2}{*}{3}
& 48.4/68.1 & 45.5/60.1 & 48.4/67.1 & {\bf 50.0}/74.3 & \bf 59.5/121.9 & \multirow{2}{*}{n/a} \\
& 45.5/61.4 & 43.6/59.4 & 46.4/63.7 & 51.1/73.9 & 62.7/141.0 \\\hline

\multirow{2}{*}{4}
& 58.3/78.7 & {\bf 54.2}/76.0 & 54.4/{\bf 78.0} & 63.8/{\bf 91.9} & 78.6/183.5 & \multirow{2}{*}{n/a} \\
& 54.2/73.7 & 55.6/73.2 & 53.0/80.2 & 62.9/92.1 & 77.9/175.6 \\\hline

\multirow{2}{*}{5}
& 65.3/90.5 & {\bf 62.6}/87.1 & 65.8/96.7 & 76.3/{\bf 117.7} & \bf 77.9/162.4 & \multirow{2}{*}{n/a} \\
& 65.3/90.1 & 64.0/86.2 & 63.1/96.0 & 75.7/118.1 & 90.4/210.5 \\\hline

\multirow{2}{*}{6}
& {\bf 75.2}/105.2 & 76.7/110.8 & \bf 72.0/101.4 & {\bf 77.0}/117.4 & 88.4/186.6 & \multirow{2}{*}{n/a} \\
& 79.5/99.8  & 74.8/110.7 & 80.8/111.6 & 81.7/115.7 & n/a \\\hline

\end{tabular}
\caption{Model sizes (top block), BLEU scores (middle block) and decoding times (bottom block) for the ASPEC \unidir{Ja}{En} task. Results are for 1- to 6-layer RS and vanilla models with hidden-filter sizes of 64-256, 128-512, 256-1,024, 512-2,048 (default), 1,024-4,096, and 2,048-8,192. In the middle and bottom blocks, each cell contains 2 pairs of 2 scores, except for identical 1-layer RS and vanilla models.  The first pair of scores is for the RS model and second one is for the vanilla model. Each pair of scores separated by a ``/'' indicates greedy search and beam-search BLEU scores or decoding speeds. Entries marked with a ``n/a'' are those for which we were unable to train and hence decode models due to lack of computational capacity (GPU memory). Numbers marked bold indicate that they are superior to the corresponding vanilla models shown in the second line.}
\label{table:costbenefitaspec}
\end{table*}

The results are shown in \Tab{costbenefitaspec}. Before delving into details, we would like to emphasize that models with larger hidden-filter sizes (1,024-4,096\footnote{Note that these hidden-filter sizes are the same as used in the Transformer Big setting. Big models are usually better than Base models in high-resource settings \cite{NIPS2017_7181}, and RS models that tend to be faster to decode than vanilla models in these situations are definitely worthwhile.} and above) tend to show better performance despite them used in shallow layer models. Furthermore, RS models barely differ in translation quality compared to vanilla models despite their reduced capacity.
With regards to translation quality, when using small hidden-filter sizes (64-256 and 128-512), vanilla models always outperform RS models, especially when shallow layers are used. However, for hidden-filter size of 128-512, the deepest 6-layer vanilla and RS models are within approximately 1.0 BLEU of each other. Given that we lose up to approximately 1.0 BLEU points for reducing 20\% parameters, we consider that 6-layer RS models with 128-512 hidden sizes can be the smallest possible viable NMT models that can be used in place of vanilla models. When using hidden-filter sizes of 256-1,024 and above, the gaps between vanilla and RS models reduce even further.
In some configurations marked bold in the table, RS models outperform corresponding vanilla models in terms of BLEU score.
In other cases, the BLEU score differences are not statistically siginificant. In these settings, RS models help reduce the number of parameters from 10\% to 50\%. We consider that this is reason enough to prefer RS models in favor of vanilla models.

When focusing on efficiency in terms of decoding speed, the number of encoder-decoder layers has a larger impact than the hidden-filter sizes in most cases. In fact, the decoding speeds for hidden-filter sizes up to 512-2,048 are mostly similar when the number of encoder-decoder layers are fixed. However, when using larger hidden-filter sizes,
the decoding times increase. Furthermore, when using smaller hidden-filter sizes, the decoding speeds between vanilla and RS models do not differ by much, but when using hidden-filter sizes larger than 512-2,048, RS models tend to be faster. We speculate that this is because larger hidden-filter sizes end up utilizing the caches and CUDA cores of the GPUs to their limit.  Thus, RS models are more effective as the amount of parameters needed for caching are fewer. This observation is in line with previous work \cite{Li2019}. Given that the RS models tend to give the better performance on average when using larger hidden-filter sizes (1,024-4,096 and above) while being slightly faster to decode than vanilla models for this ASPEC \unidir{Ja}{En} task, we consider that RS models are certainly a viable alternative to vanilla models. In practice, we recommend an in-depth exploration like we did before choosing a model that satisfies all cost-benefit requirements.

While we were unable to train vanilla models with 2 or more layers and hidden-filter sizes of 2,048-8,192 due to our GPU infrastructure, our observations should be generalizable to these and even larger hidden-filter size settings. We performed this cost-benefit analysis for the GCP \unidir{En}{Ja} task as well. Our observations related to decoding speeds were mostly similar to when using wide hidden sizes (1,024 or 2,048) for the ASPEC \unidir{Ja}{En} task. Similarly, the large gaps in translation quality for smaller hidden sizes (up to 128) were also observed. However, the vanilla model tended to give slightly better translations than RS models regardless of the model size setting. We would like the readers to note an important point: the GCP tasks are relatively low-resource compared to ASPEC, and using wider layers does involve the possibility of over-fitting and this makes it possible for the scores to be rather unreliable unless additional analysis involving regularization is done. Nevertheless, we have shown that there exist settings (corpora sizes and model sizes) where a vanilla model can certainly be replaced by an RS model.

\begin{figure*}[t]
    \centering
    \begin{subfigure}[t]{0.48\textwidth}
    \includegraphics*[width=\textwidth]{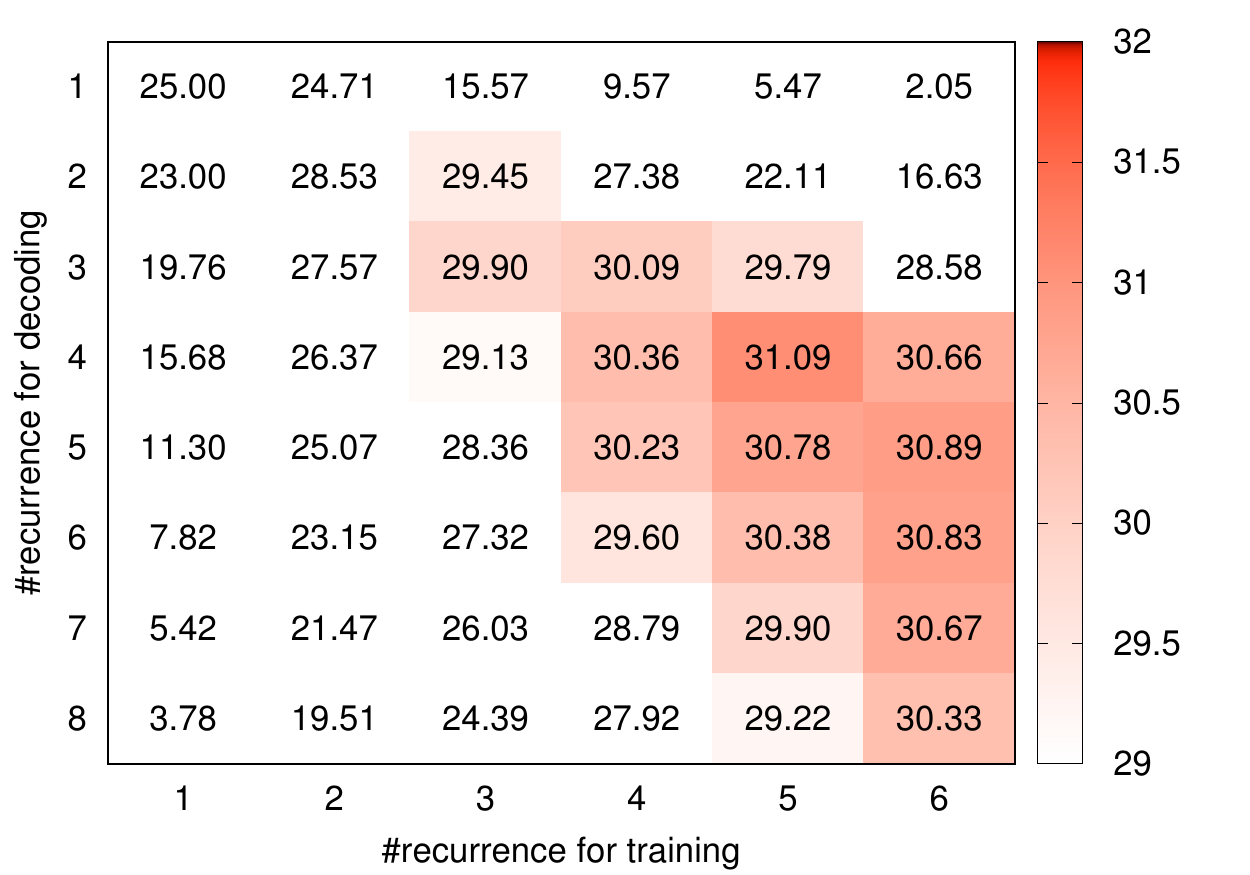}
    \caption*{(b)~RS-NMT scratch.}
    \end{subfigure}
    \begin{subfigure}[t]{0.48\textwidth}
    \includegraphics*[width=\textwidth]{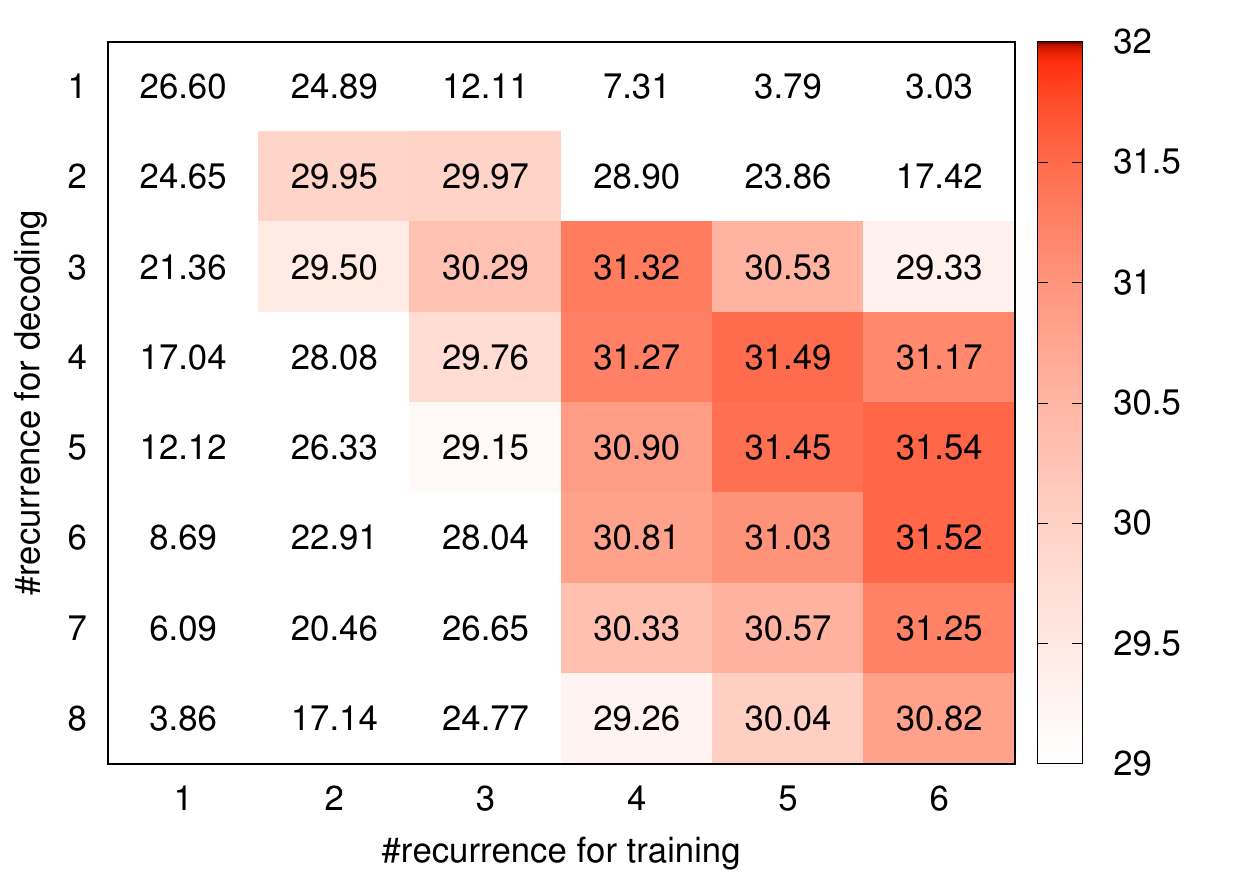}
    \caption*{(d)~RS-NMT distilled.}
    \end{subfigure}
    \caption{BLEU scores of the RS-NMT models for the KFTT \unidir{En}{Ja} task with a variable number of recurrences of decoder layers during decoding. The number of encoder recurrences is the same as during training.}
    \label{figure:recurrent-decoding}
\end{figure*}

\section{Analyzing Recurrently Stacked NMT Models}
\label{section:analysis}

The above results and analyses show that RS models are best trained with transfer learning methods leading to compact and fast models which should be extremely useful in real-time resource-constrained and latency-sensitive scenarios. We now focus on understanding the internal working of RS models through analyzing their decoding behavior and visualizing parts of the models inner components.
In particular, we answer the following two questions.
\begin{description}
    \item[(a)] Do RS-NMT models memorize the number of recurrence?
    \item[(b)] Do RS-NMT models behave differently from vanilla NMT models internally?
\end{description}

\subsection{Number of Recurrence Memorized by RS-NMT}

Any deep stacked model trained with $N$ layers can theoretically be used for decoding with fewer than $N$ layers. Non-RS models cannot be used with more than $N$ layers, because the parameters for deeper layers do not exist. In contrast, RS models have the same parameters regardless of the number of layers and recurrence and thus can be used for decoding more flexibly with fewer or more layers. As we have seen in \Tab{direct-results}, the performance of RS-NMT models seems to stabilize with the increasing number of RS layers. As such, we expected that the representations obtained by deep RS should be robust and thus enable the use of fewer layers during decoding than those used during training. To confirm whether this holds or not, we trained several RS-NMT models and used them for decoding with different times of recurrence, taking the KFTT \unidir{En}{Ja} task as an example.

When performing decoding, we fixed\footnote{In our previous work \cite{DBLP:conf/aaai/DabreF19}, we have shown that using the same number of recurrences between training and decoding is important for the encoder. Reducing the number of encoder layers does not have a large impact, because the encoding process of the Transformer architecture is highly parallelizable. In contrast, reduction of the number of decoder layers is valuable, because its auto-regressive process is time-consuming.} the number of recurrences for the encoder to that what was used during training, and varied only the number of recurrences for the decoder from 1 to 8. For example, a trained 6-layer RS-NMT model was tested with 6 times of RS for the encoder (same as training) but do 1 to 8 times of RS for the decoder. We evaluated two types of RS-NMT models: ones trained from scratch and the others trained via sequence distillation. We did not do this for the fine-tuning models for two reasons.  First, there are a number of fine-tuned models which would be difficult to analyze.  Second, fine-tuned models are trained using the same data as those trained from scratch and hence there is no reason to expect any fundamental change in their behavior regarding recurrence memorization.

\newcommand{\exone}{{\textit{Ippan'ni wa Dogen Zenji to yoba reru .}}}
\newcommand{\trone}{{He is generally called Dogen Zenji .}}

\begin{figure*}[t]
    \centering
    \begin{subfigure}[b]{\textwidth}
    \includegraphics*[width=1.0\hsize]{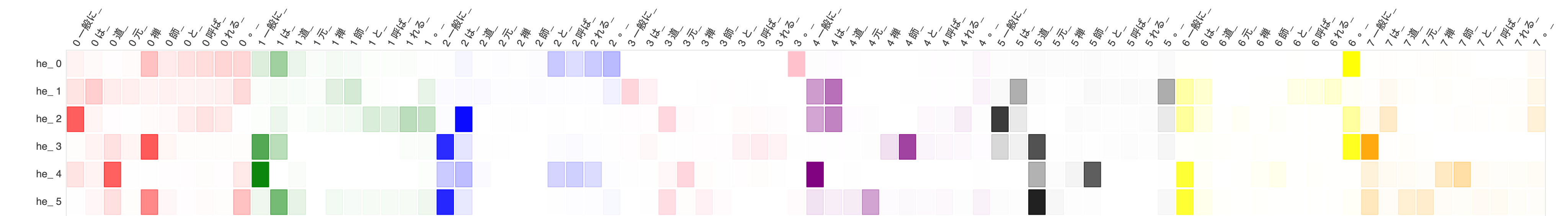}
    \caption{Vanilla NMT.}
    \end{subfigure}
    \begin{subfigure}[b]{\textwidth}
    \includegraphics*[width=1.0\hsize]{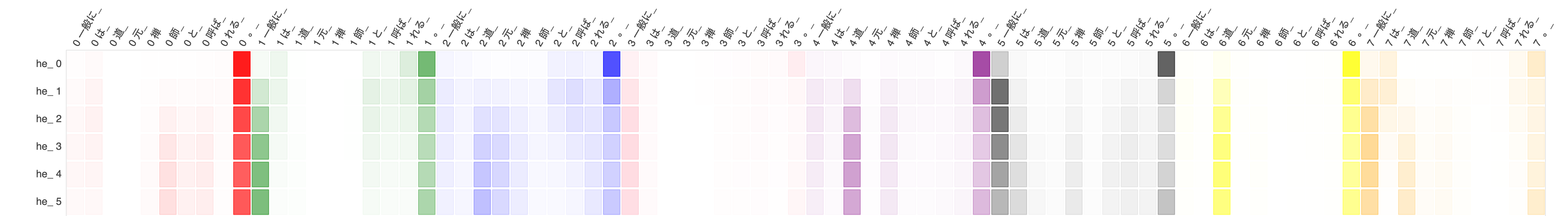}
    \caption{RS-NMT scratch.}
    \end{subfigure}
    \begin{subfigure}[b]{\textwidth}
    \includegraphics*[width=1.0\hsize]{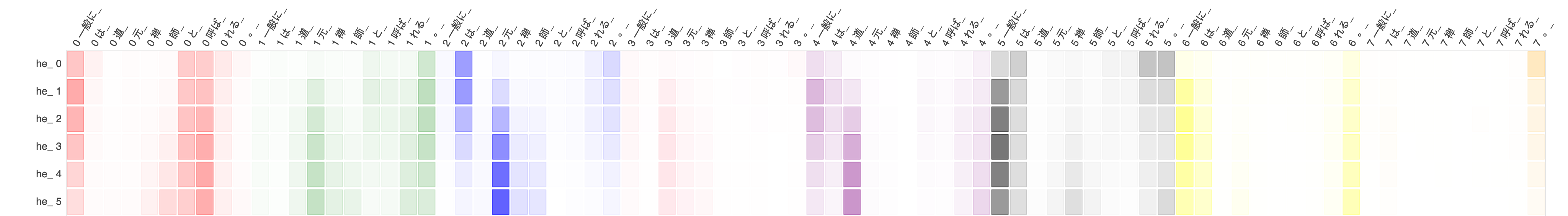}
    \caption{The best RS-NMT using fine-tuning (initialized by the 3rd layer of the vanilla NMT model).}
    \end{subfigure}
    \begin{subfigure}[b]{\textwidth}
    \includegraphics*[width=1.0\hsize]{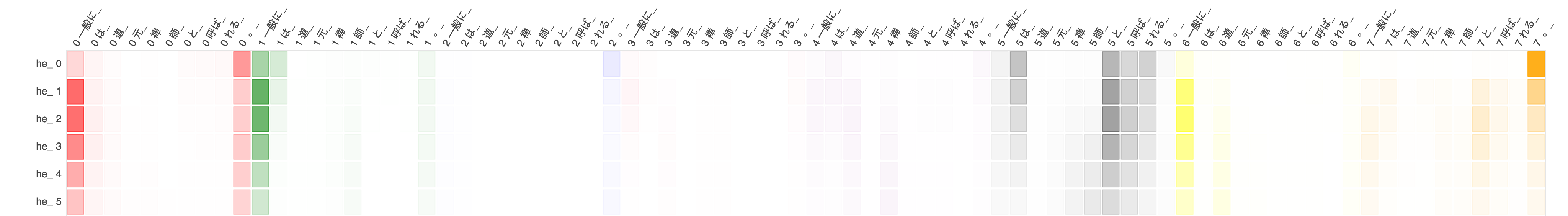}
    \caption{RS-NMT distilled.}
    \end{subfigure}
    \caption{Visualization of the cross-attention of the four different types of NMT models trained for the KFTT \unidir{Ja}{En} task. This visualization shows only a single sub-word out of the entire sequence in the target language for an input sentence, ``\exone'' (\trone). The x-axis indicates the input sentence, i.e., sequence of sub-words, for each of the 8 attention heads (marked by different prefix of each sub-word and different colors), whereas the y-axis indicates the 6 layers with 0-to-5 in suffix representing 1st-to-6th layer from top to the bottom, respectively. Each cell shows the cross-attention probability given by a head at a layer to a source sub-word, when generating the target sub-word, where darker color means stronger attention with higher probability.}
    \label{figure:attvis}
\end{figure*}

\Fig{recurrent-decoding} summarizes the BLEU scores achieved by the above two types of models.
Irrespective of the use of sequence distillation, regarding each model shown in a column of the figures, the number of recurrences during training is not precisely memorized, and the deeper models have a higher level of flexibility. For example, the 6-layer models used with 4--7 times of recurrences achieved comparable BLEU scores.  In contrast, there is no guarantee about the performance with fewer number of recurrences, as evidenced by the sharp drop of BLEU score when used with 1--3 decoder layers.  BLEU scores also dropped when we used the models with more recurrences than what had been used for training.  These observations indicate that the computation of the most useful and hence the most reliable features takes place around the deepest layer used during training.
As such, we conclude that the RS-NMT, in its current form, is unable to generalize the variation in the number of recurrence between training and decoding. However, we can say that the RS-NMT models can weakly memorize the number of recurrences.

Note that sequence distillation yielded better RS-NMT models than those trained from scratch. To be even more precise, the 4-layer RS-NMT model obtained via sequence distillation was as good as or better than the best RS-NMT model (6-layer) trained from scratch. As such, we can avoid training deeper models if we use sequence distillation, and this can also save decoding time.

\subsection{Visualizing Recurrently Stacked Models}

To acquire a deeper understanding of what happens in RS-NMT models, we visualize the attentions across all attention heads for each stacked layer. In particular, we consider all the types of models trained for the KFTT \unidir{Ja}{En} task, except vanilla-vanilla distilled one, and visualize their encoder's self-attentions, the decoder's self-attentions, and the decoder's cross-attentions.  For the sake of simplicity, we only show the visualizations for cross-attentions of a target sub-word with the entire source sentence.  See \App{visualization} for the sentence-level cross-attention visualizations for the same example.

\Fig{attvis} displays the heat-maps for the 8-head encoder-decoder cross-attention across all 6 layers
when generating the first translated sub-word, ``The,'' for an input Japanese sentence consisting of eight tokens (ten sub-words), ``\exone'' (\trone).
Note that the NMT models have eight attention heads that are shared across 6 RS layers, whereas the vanilla model has 48 (8$\times$6) different attention heads.
As shown in the figure, the attention mechanism behaves differently for the vanilla NMT and RS-NMT models, whereas there is no noticeable difference  among the three RS-NMT models themselves. There is no consistency in the sharpness of attention as we go deeper in the vanilla NMT model.  In contrast, as we go deeper in the RS-NMT models, the attention tends to stabilize itself. To be more precise, it seems to be stable around the 3rd or 4th layer and barely change in deeper layers. This could be a possible explanation as to why the BLEU scores do not vary by large amounts despite using a variable number of recurrences during decoding as compared to training, as we have seen in \Fig{recurrent-decoding}.

\begin{figure}[t]
    \centering
    \begin{subfigure}[t]{0.4\columnwidth}
    \includegraphics*[width=.9\columnwidth]{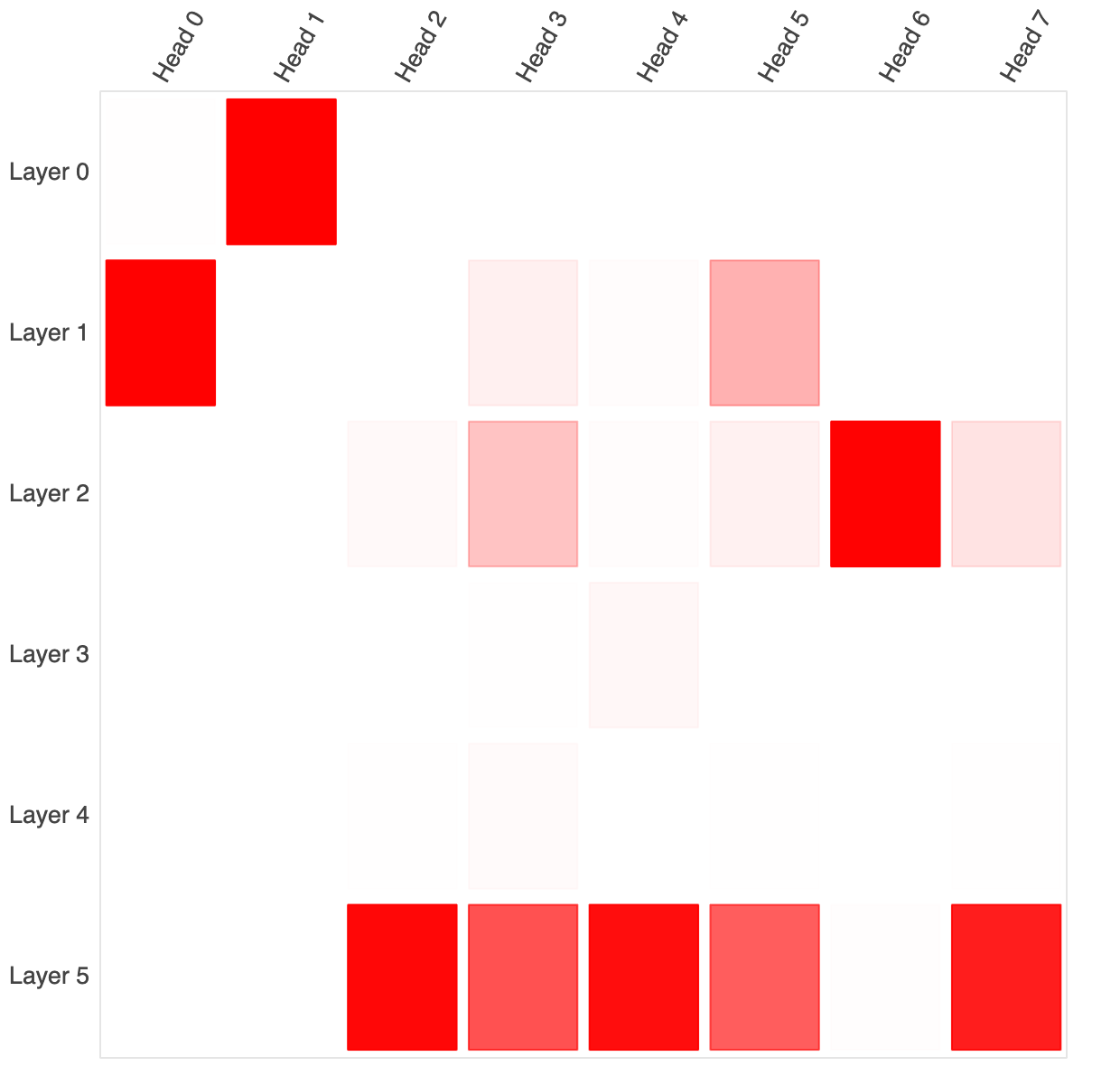}
    \caption{Vanilla NMT.}
    \end{subfigure}
    \qquad
    \begin{subfigure}[t]{0.4\columnwidth}
    \includegraphics*[width=.9\columnwidth]{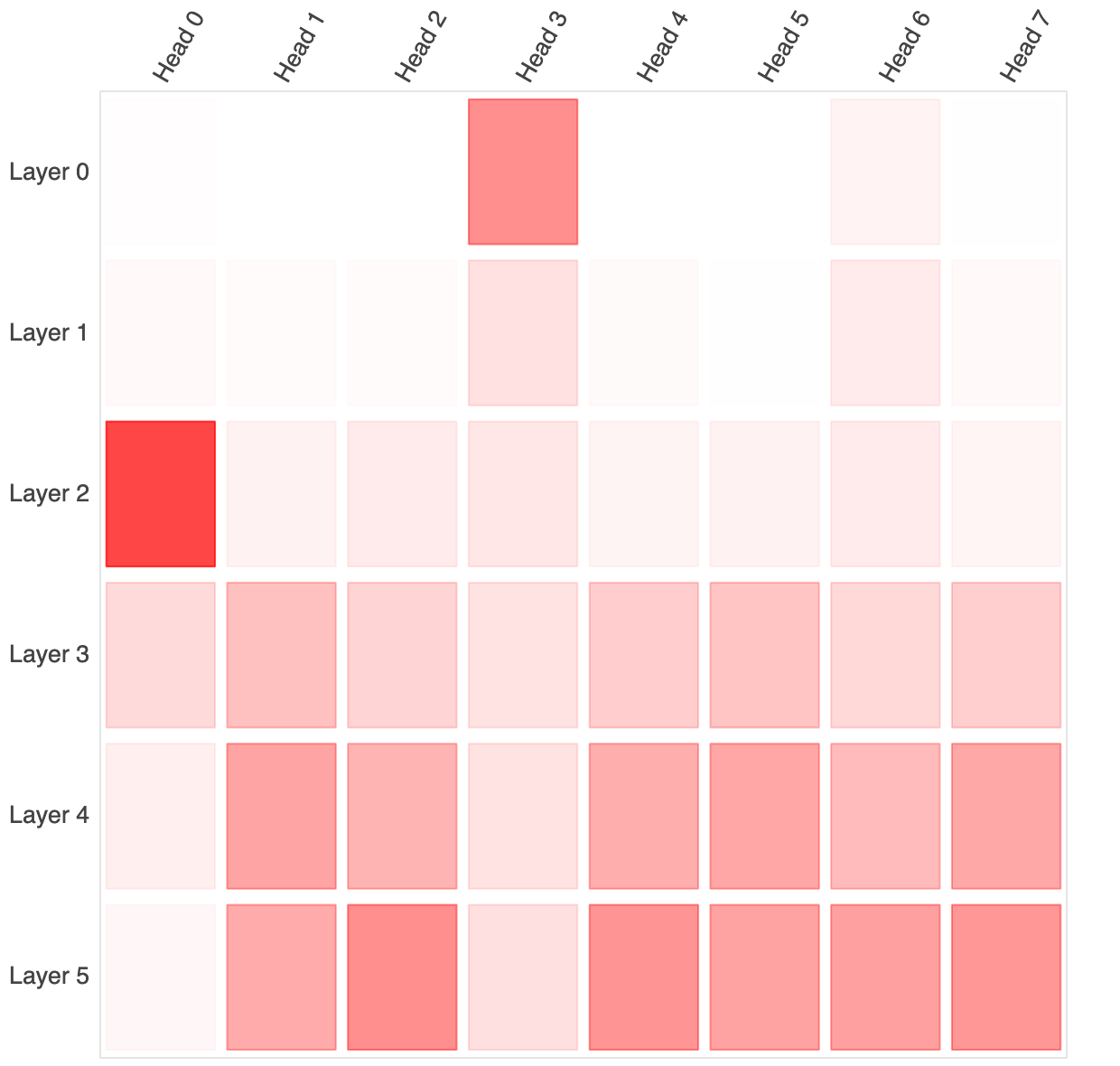}
    \caption{RS-NMT scratch.}
    \end{subfigure}
    \\
    \begin{subfigure}[t]{0.4\columnwidth}
    \includegraphics*[width=.9\columnwidth]{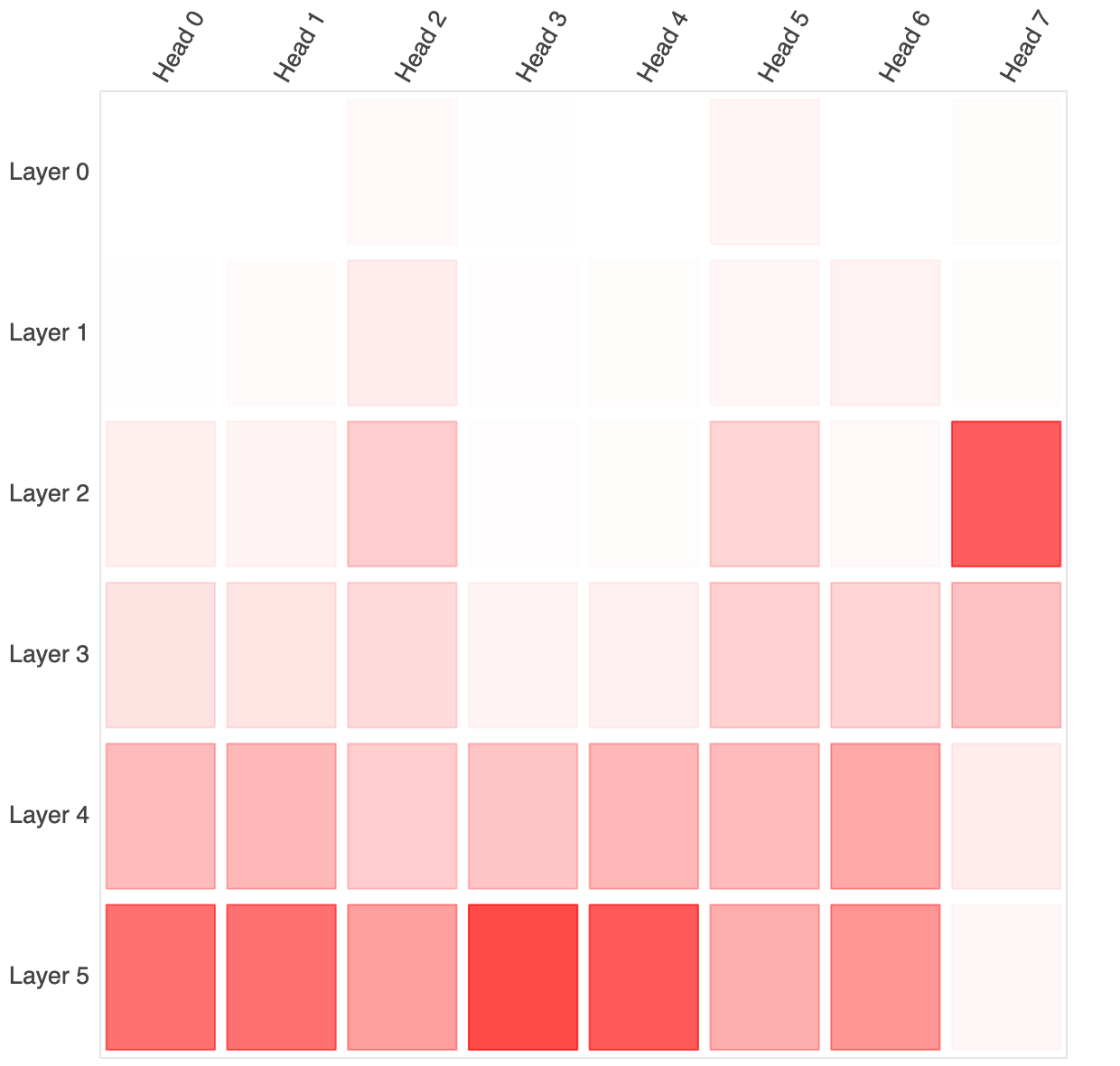}
    \caption{The best RS-NMT using fine-tuning (initialized by the 3rd layer of the vanilla NMT model).}
    \end{subfigure}
    \qquad
    \begin{subfigure}[t]{0.4\columnwidth}
    \includegraphics*[width=.9\columnwidth]{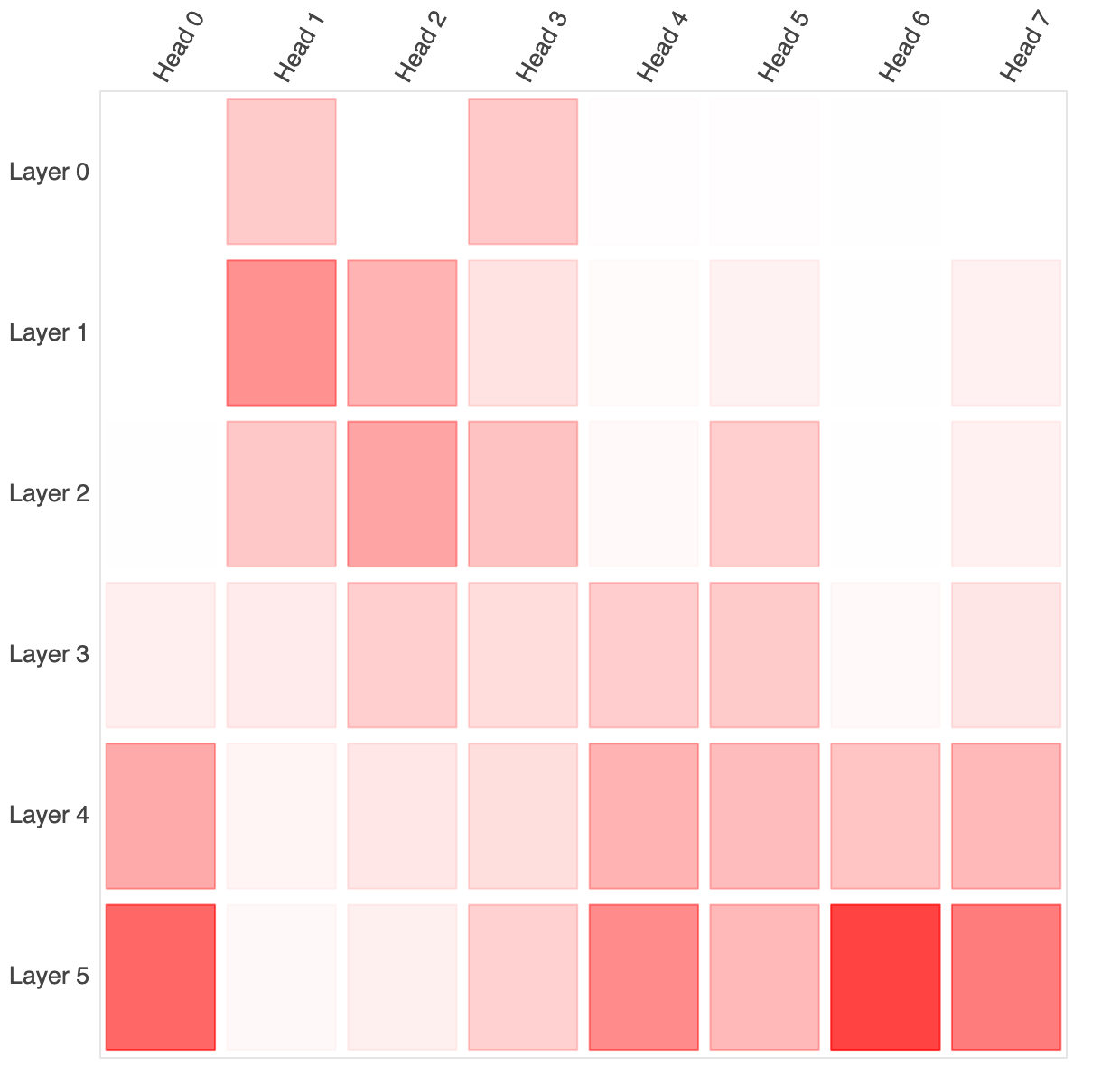}
    \caption{RS-NMT distilled.}
    \end{subfigure}
    \caption{Visualization of attention entropies of the four different types of NMT models (same as in \Fig{attvis}) for the same input sentence, ``\mbox{\exone}'' (\trone). The x-axis indicates the attention heads, whereas the y-axis indicates the depth of the layer (top: 1st layer, bottom: 6th layer).  A darker color means higher value of entropy and more uniform distribution of attention probability.}
    \label{figure:entrvis}
\end{figure}

For a further understanding, in \Fig{entrvis}, we also produced heat-maps each of which displays for each model the entropies of the attentions for each attention head in each layer.
Higher entropy (darker color) means that the attention head focuses on more sub-words, whereas lower entropy (lighter color) indicates that the head attends to a fewer number of sub-words.
In the vanilla NMT model, the entropies of the attention heads are often very close to zero, as a result of intensive attention to a single sub-word, such as in the 3rd head of the 3rd layer (Head 2 of Layer 2 in \Fig{entrvis}).  Otherwise, the entropy value can be very high, indicating that the head attends to almost all the sub-words nearly uniformly, as seen in the 1st head of the 2nd layer (Head 0 of Layer 1 in \Fig{entrvis}).  We could not observe any layer-wise tendency, presumably owing to the independence nature of attention heads.
In contrast, in the RS-NMT models, the attention entropy of more than half of the attention heads is non-zero, meaning that each of these heads attends to more than one sub-words, but at the same time they are more conservative by focusing on fewer sub-words than the vanilla model.
It is noteworthy that the attention entropy tends to gradually stabilize as we move towards deeper layers, which has also been seen in \Fig{attvis}.

In the RS-NMT models, both the attention values and the entropies seem to calibrate themselves. This means that the attention mechanism somehow learns to adjust itself and dynamically considers the contribution from more or fewer sub-words by giving them similar or dissimilar weights. More often than not, the entropies tend to grow which means that deeper layers prefer to focus on more sub-words. This ability to self-calibrate possibly comes from the fact that the same parameters are used across all RS layers, meaning that all computed representations will lie in the same vector space. In contrast, the parameters of the attention heads across layers in vanilla NMT models are initialized independently and hence all computed representations will lie in separate vector spaces. Consider the work on average attention networks \citep{DBLP:conf/acl/XiongZS18} where the self-attention weights are statically set as $1/L$ where $L$ is the length of the sequence against which the attention is computed when generating a target sub-word. Average attention has the highest entropy because each word receives the same amount of attention. As RS models tend to have higher entropies for deeper layers, we suppose that the attention mechanism calibrates itself to attend to more words as the depth increases. This also means that attending equally to more words is beneficial. As such, our observation for RS models, that attend to more words equally, might be an explanation of why average attention networks, which attend to all words equally, work well despite using an ordinary attention mechanism.

We can also relate this to the principle of maximum entropy\footnote{\url{https://en.wikipedia.org/wiki/Principle_of_maximum_entropy}} which states that the probability distribution that best represents the data is the one with the maximum entropy. Given that RS models tend to have higher attention entropy across all heads in the deeper layers, we hypothesize that the attention models are indirectly trained to satisfy a maximum entropy objective in order to better represent the context used during decoding. We will explore this hypothesis in the future.

Another probable explanation lies in the understanding of what recurrent neural networks (RNNs), such as long short-term memories (LSTMs) \citep{Hochreiter:1997:LSM:1246443.1246450}, do. At each time-step, an RNN refines the representation of a sequence as it processes each new sub-word with the same parameters. As such, we speculate that the deeper layers of an RS-NMT model are forced to learn more abstract representations through refining those in the shallower layers. The parameter sharing, and thus the reduction in representational power, most likely causes all artificial neurons to work in unison and uniformly distributes the labor of generating reliable representations. This would mean that the learned representation for each sub-word is rather generic. And this in turn causes the attention mechanism to seek out more work for the decoder to compute reliable representations.

Since the attentions in RS-NMT and vanilla NMT models differ greatly, they might have different applications. For example, RS-NMT attentions might be more suited for phrase-to-phrase matching due to the wide attention spans it tends to exhibit. It would be worthwhile to visualize the word- and sentence-level representations to explore our hypotheses.

\section{Conclusion}
\label{section:conclusion}

In this paper, aiming to obtain a compact neural network model, we have proposed recurrent stacking (RS) of layers.
In the neural machine translation (NMT) tasks, we confirmed that the RS-NMT models trained directly on the given parallel data consistently achieve much higher BLEU scores than the identical 1-layer vanilla models, whereas they still underperform the 6-layer vanilla models.  Nevertheless, despite the small number of parameters equivalent to that of the 1-layer vanilla models, their performance can approach or even surpass that of the 6-layer vanilla models when trained through transfer learning, such as knowledge distillation and/or layer transfer followed by fine-tuning. Furthermore, we showed that transfer learning approaches help reduce decoding time by reducing the need for beam search and/or deep layers. An important observation we made is that in high-resource settings, shallow RS models using wide hidden layers (512 or 1,024) trained via transfer learning tend to outperform vanilla models (also trained via transfer learning) in terms of translation quality, model size and, in some cases, decoding speed.

Through varying the number of the decoder layers of the RS-NMT models, we show that the number of recurrence used for training is weakly memorized by the trained model itself and it is best to use the same number of recurrences during training and decoding, whereas deep RS models could be used with slightly fewer numbers of decoder layer. Our analysis of the attention mechanism also reveals that the parameter sharing by RS leads to a self-calibrating attention behavior: the attention values tend to converge as we go deeper. Furthermore, more heads are involved in attending to the source and target sub-words as evidenced by higher entropies in deeper layers indicating intensive use of the few parameters in the RS models.

The following methods can augment our proposed method and help in either improving the translation quality or compressing the model further.
\begin{description}
\item[Extremely deep and compact models:] Vanilla models that do not employ any parameter sharing between layers suffer from the risk of parameter explosion. RS models can approach the performance of multi-layer vanilla models, while retaining the number of parameters equivalent to that of 1-layer vanilla models. In our experiments, however, RS-NMT models have reached their performance limit at 6 layers similarly to vanilla NMT models. This is presumably because of so-called gradient flow issues \cite{bapna-etal-2018-training}. Advances in training deeper models should help train extremely deep RS models with an improved performance. It is also possible that the representations of RS layers converge, and mathematical analyses of such a convergence behavior should give us useful insights.
\item[Flexible Decoding of NMT models:] Our analysis has revealed that a trained RS-NMT model can be used for decoding with the fewer layers than during training without an appreciable loss.  It will be useful to have a method to train NMT models which can be used with any number of layers without a sharp loss in translation quality that existing models suffer from. Such a method should be useful for both vanilla NMT and RS-NMT models. Note that flexible decoding can have larger impact if we can combine it with the insights gained from the study of convergence behavior of RS layers.
\item[Limits of sharing parameters:] Concurrent studies have shown that sharing the self-attention and feed-forward layer parameters between the encoder and decoder is possible without a great loss in performance \cite{DBLP:conf/aaai/XiaHTTHQ19}. However, its combination with RS perform badly. Sequence distillation should help mitigate this problem. Additionally, it is unclear how  multilingual models that rely on parameter sharing will behave when combined with RS. The ultimate question in this direction is to find additional parts of neural network models to share.
\item[Filling the capacity with external data:] Using back-translation is a prominent way to improve the decoder using the monolingual data of the target language \citep{sennrich-haddow-birch:2016:P16-11}, and we have proven its effectiveness in improving RS-NMT models, indicating that RS-NMT models can afford to manage more information. We have also shown that RS-NMT benefits from sequence-level knowledge distillation and can achieve comparable results to vanilla NMT models. This is another indicator of its ability to incorporate more information. Forward translation and filtering \citep{ueffing:07} is a type of knowledge distillation and has been proven effective in improving translation performance in the form of self-learning. It should be possible to combine both these methods to incorporate into RS-NMT models, more information of source and target languages in addition to the limited parallel data.
\end{description}

\section*{Acknowledgments}

A part of this work was conducted under the program ``Research and Development of Enhanced Multilingual and Multipurpose Speech Translation System'' of the Ministry of Internal Affairs and Communications (MIC), Japan.

\bibliographystyle{acl_natbib}
\bibliography{aaai19}

\appendix
\newpage
\section{Complete Attention Visualizations}
\label{appendix:visualization}

\begin{figure*}[hb]
      \centering
    \includegraphics*[width=1.0\hsize]{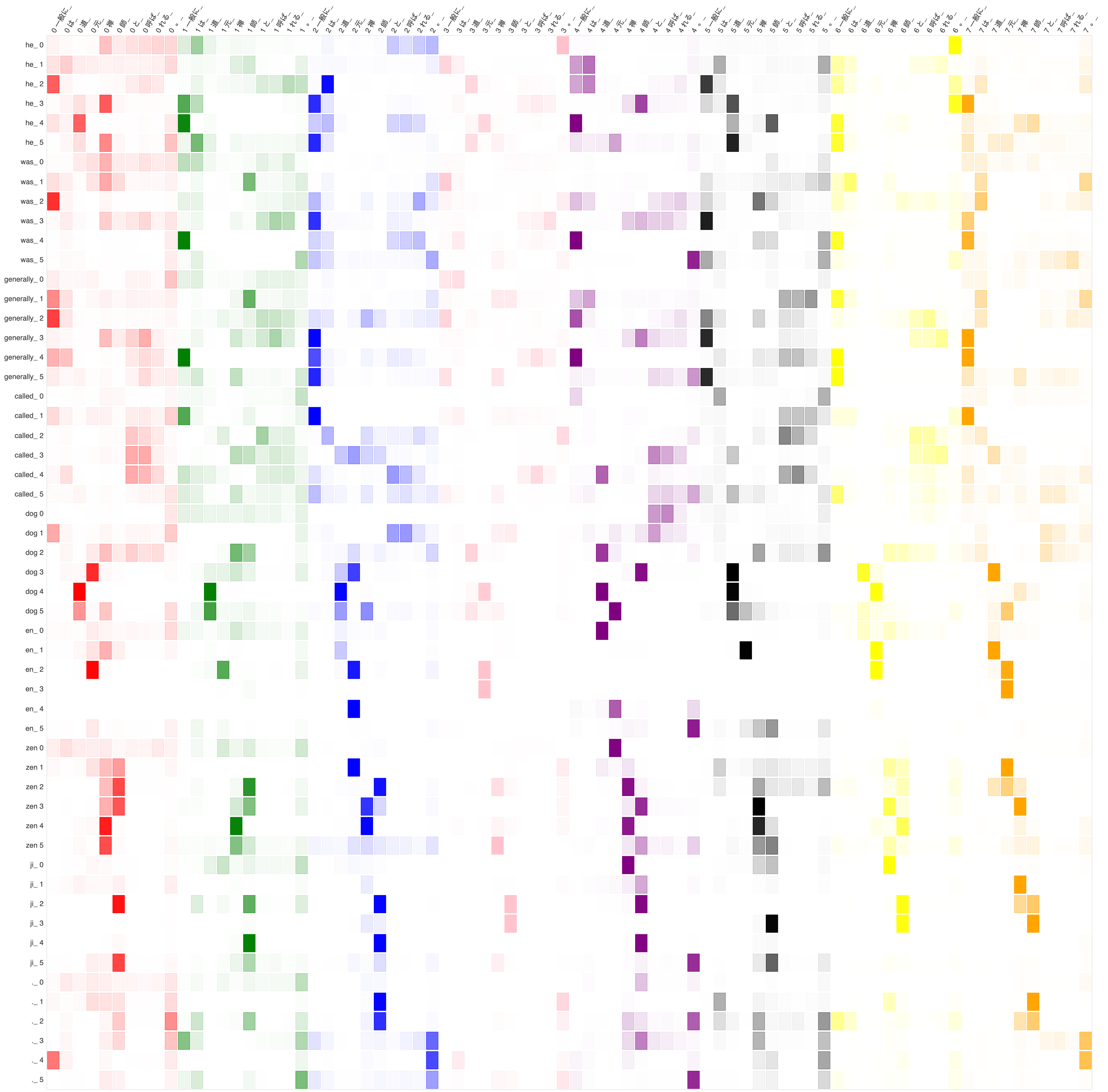}

\caption{The attention heat-map for the cross-attention of the vanilla NMT model for the KFTT \unidir{Ja}{En} task for the input sentence ``\exone'' (\trone).}
        \label{figure:vanattfull}
\end{figure*}

\begin{figure*}[ht]
      \centering
    \includegraphics*[width=1.0\hsize]{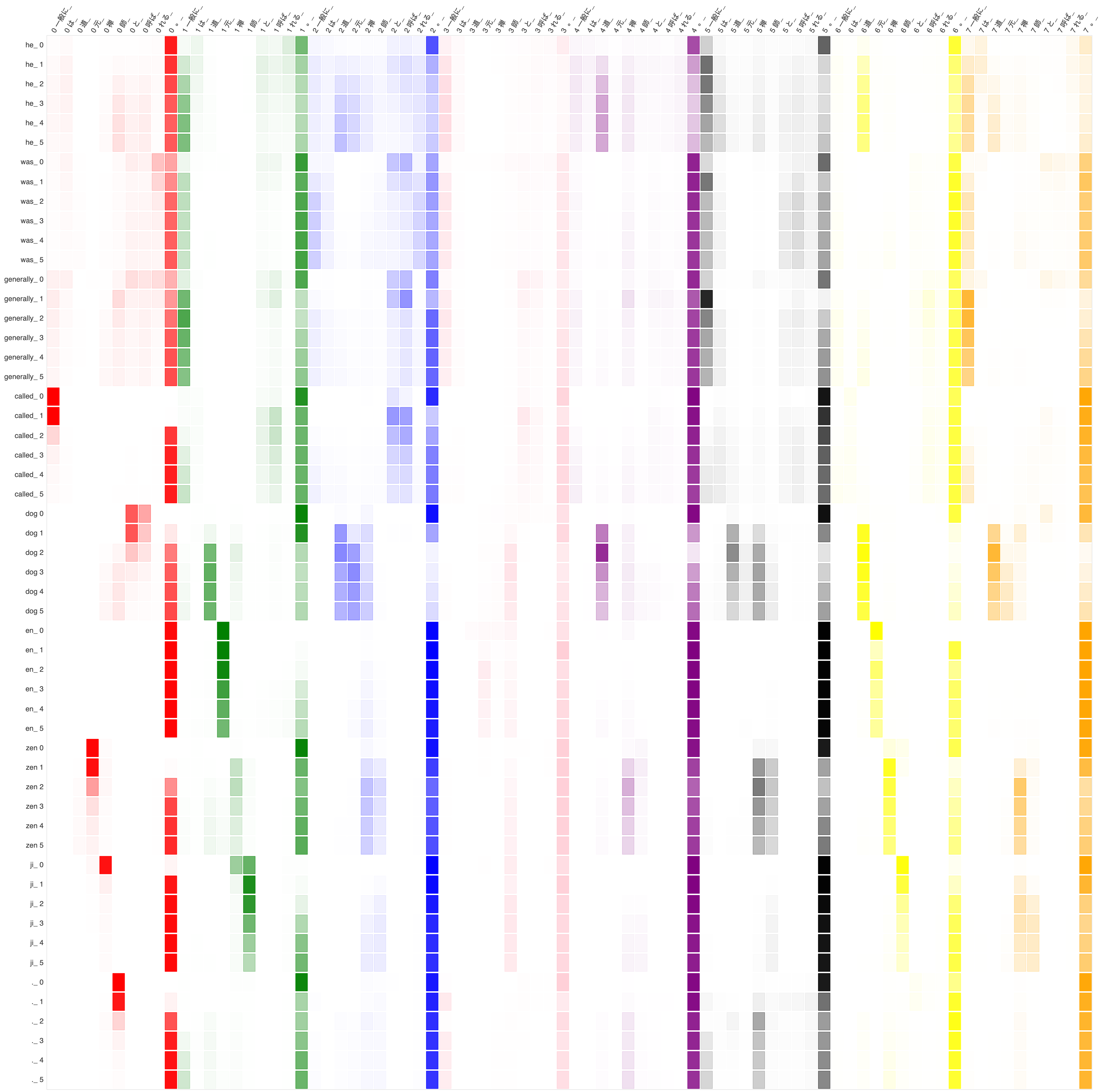}

\caption{The attention heat-map for the cross-attention of the RS-NMT model for the KFTT \unidir{Ja}{En} task for the input sentence ``\exone'' (\trone).}
        \label{figure:reccattfull}
\end{figure*}

Figures~\ref{figure:vanattfull} and \ref{figure:reccattfull} give the attention heat-maps of the cross-attention for the vanilla and RS-NMT models, respectively. The input sentence for both models is ``\exone'' (\trone) and the models are trained for the KFTT \unidir{Ja}{En} task. While there seems to be no agreement between the attention heads in the vanilla NMT model across layers, the attention heads in the RS-NMT model seem to find fixed parts of the sentence to attend to and then calibrate the amount of attention as the depth increases. Note that the translations for the same input sentence can be different depending on the model chosen.

\end{document}